\documentclass[]{bytedance_seed}

\usepackage[toc,page,header]{appendix}

\usepackage{minitoc}
\usepackage{booktabs}
\usepackage{makecell}

\newcommand{\ours}{\textit{PHD}}

\newcommand{\oursswa}{\textit{PHD-SWA}}

\newcommand{\ourscswa}{\textit{PHD-CSWA}}

\title{ Efficient Pretraining Length Scaling }

\author[1, \dagger]{Bohong Wu}
\author[1,2]{Shen Yan}
\author[1]{Sijun Zhang}
\author[1,3]{Jianqiao Lu}
\author[1]{Yutao Zeng}
\author[1]{Ya Wang}
\author[1]{Xun Zhou}

\affiliation[1]{ByteDance Seed}
\affiliation[2]{Peking University}
\affiliation[3]{Hong Kong University}

\contribution[*]{Work done at ByteDance Seed}
\contribution[\dagger]{Corresponding authors}

\abstract{
Recent advances in large language models have demonstrated the effectiveness of length scaling during post-training, yet its potential in pre-training remains underexplored.
We present the Parallel Hidden Decoding Transformer (\textit{PHD}-Transformer), a novel framework that enables efficient length scaling during pre-training while maintaining inference efficiency.
\textit{PHD}-Transformer achieves this through an innovative KV cache management strategy that distinguishes between \textit{original tokens} and \textit{hidden decoding tokens}.
By retaining only the KV cache of \textit{original tokens} for long-range dependencies while immediately discarding \textit{hidden decoding tokens} after use, our approach maintains the same KV cache size as the vanilla transformer while enabling effective length scaling.
To further enhance performance, we introduce two optimized variants: \textit{PHD-SWA} employs sliding window attention to preserve local dependencies, while \textit{PHD-CSWA} implements chunk-wise sliding window attention to eliminate linear growth in pre-filling time.
Extensive experiments demonstrate consistent improvements across multiple benchmarks.

}

\date{\today}
\correspondence{Bohong Wu at \email{bohongwu@bytedance.com}, Xun Zhou at \email{zhouxun@bytedance.com}}

\begin{document}
\maketitle


\section{Introduction}

Recent years have witnessed the breakthrough of large language models (LLMs)~\cite{team2023gemini,team2024gemini,liu2024deepseek,liu2024deepseekv3,li2025minimax,jiang2023mistral7b} across various domains~\cite{wang2024openhands,han2024psydial,jimenez2023swe,chen2024step}. Apart from the early success of scaling in parameters and training data~\cite{geiping2025scaling,kaplan2020scaling,clark2022unified,hernandez2021scaling} in the pre-training stage, the recent success of Deepseek-R1~\cite{guo2025deepseek} and OpenAI-o1/o3~\cite{o1,o3} have stimulated researches on length scaling during the post-training stage. 
By employing RL methods including PPO~\cite{schulman2017proximal} or GRPO~\cite{shao2024deepseekmath}, model learns length scaling by generating very long chain-of-thought (COT)~\cite{wei2022chain} sequences before giving the final answers, leading to remarkable improvement on Olympiad-level math and reasoning problems including American Invitational Mathematics Examination (AIME) and GPQA~\cite{rein2024gpqa}. 

Given the great success of length scaling in post-training,
researchers have also been studying length scaling in the pre-training stage.
Early work on chain-of-thought reasoning~\cite{wei2022chain} inspires methods that insert plain text into pre-training sequences either through manual design~\cite{goyalthink} or online exploration~\cite{zelikman2024quiet}.
Recently, under the concept of converting discrete signals into continuous representations, Coconut~\cite{hao2024training} proposes inserting latent embeddings rather than plain text.
CoTFormer~\citep{mohtashami2023cotformer} achieves an implicit 2$\times$ pre-training length scaling by reusing hidden embeddings from earlier layers for a single token.
In contrast,
COCOMix~\citep{tack2025llm} emphasizes the interpretability of intermediate hidden states and projects them into continuous concepts.


Although these pioneer researches are proven effective, they have limited applicability, as their marginal improvements in reasoning benchmarks are achieved at the cost of increased KV cache size and higher decoding latency. Moreover, distracted by post-processing the middle layer hidden states in various ways, the innate pre-training length scaling phenomenon is never explored.




\begin{figure}[tp]
    \centering
    \begin{subfigure}[b]{0.23\textwidth}
        \centering
        \includegraphics[width=\textwidth]{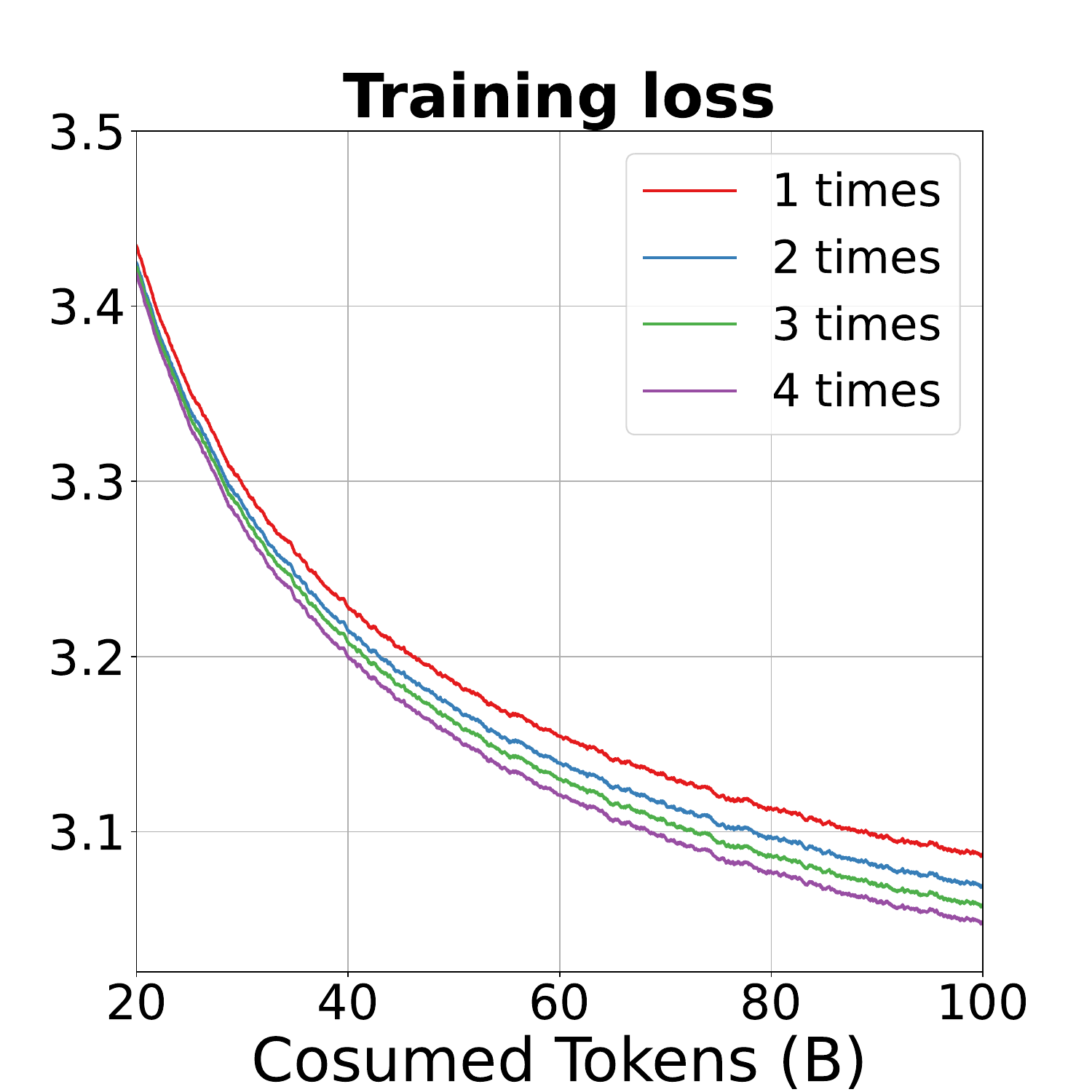}
        \caption{Training loss.}
        \label{fig:151m_trainloss}
    \end{subfigure}
    \hspace{0.01\textwidth}
    \begin{subfigure}[b]{0.23\textwidth}
        \centering
        \includegraphics[width=\textwidth]{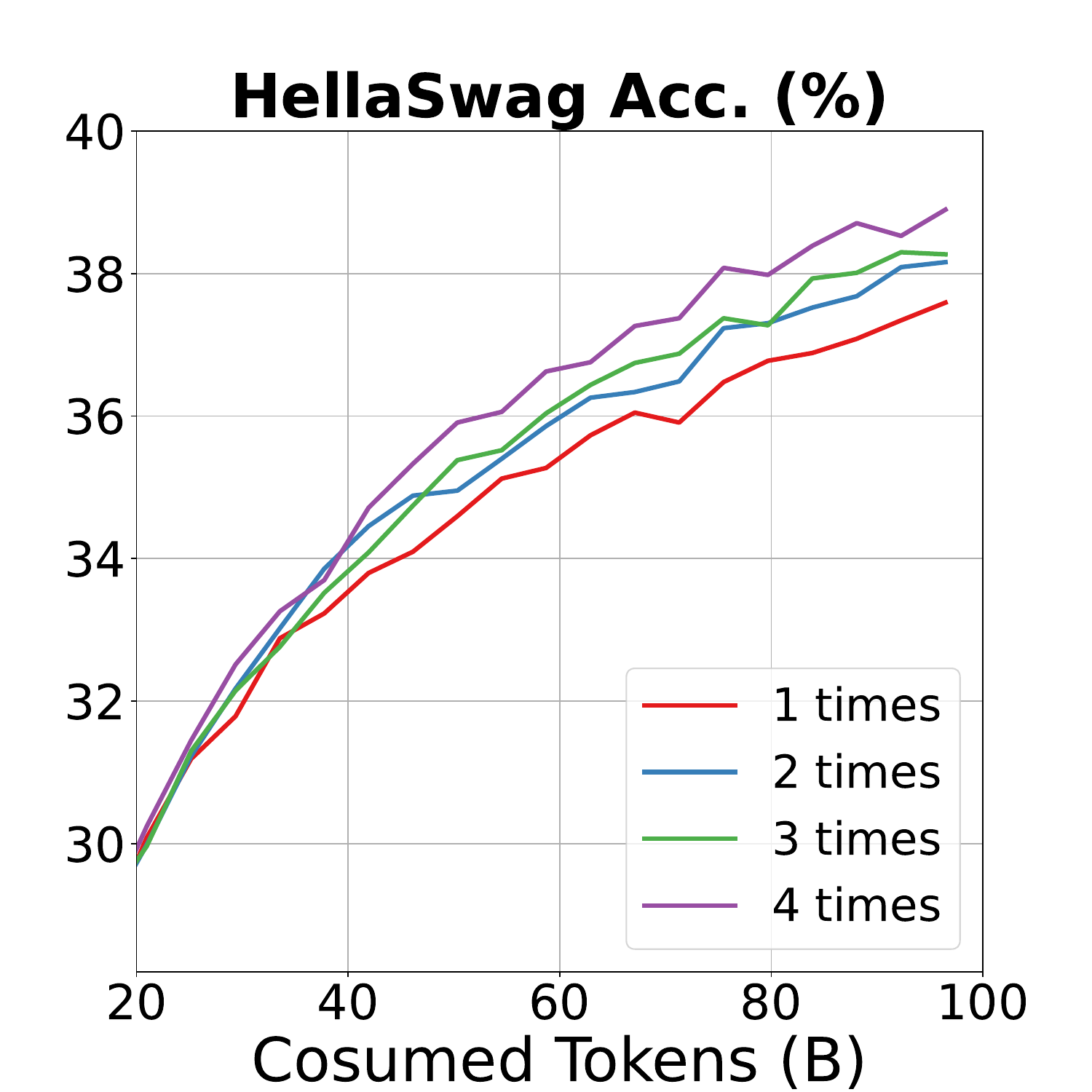}
        \caption{HellaSwag accuracy.}
        \label{fig:151m_hella}
    \end{subfigure}
    \hspace{0.01\textwidth}
    \begin{subfigure}[b]{0.23\textwidth}
        \centering
        \includegraphics[width=\textwidth]{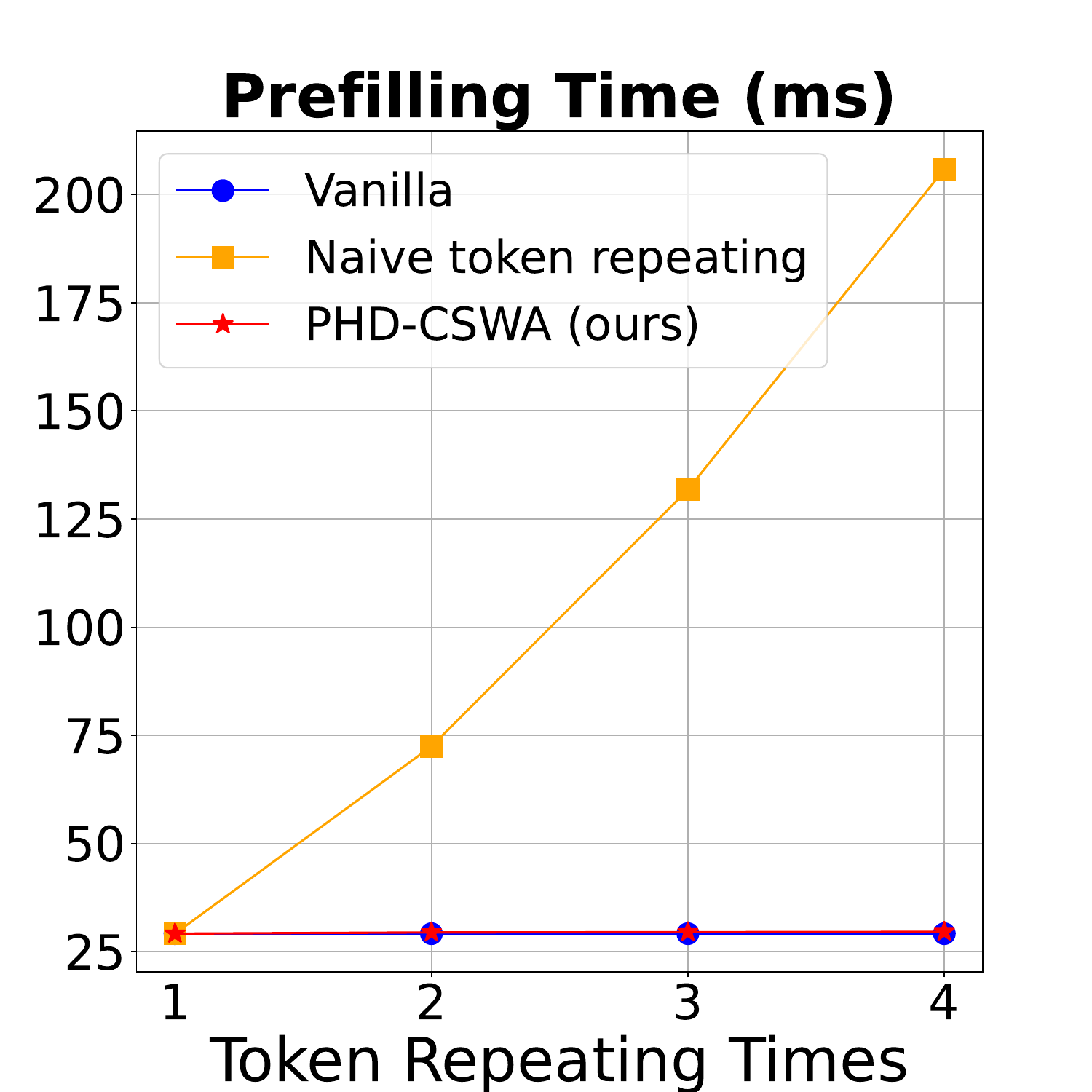}
        \caption{Prefilling time.}
        \label{fig:151m_prefill}
    \end{subfigure}
    \hspace{0.01\textwidth}
    \begin{subfigure}[b]{0.23\textwidth}
        \centering
        \includegraphics[width=\textwidth]{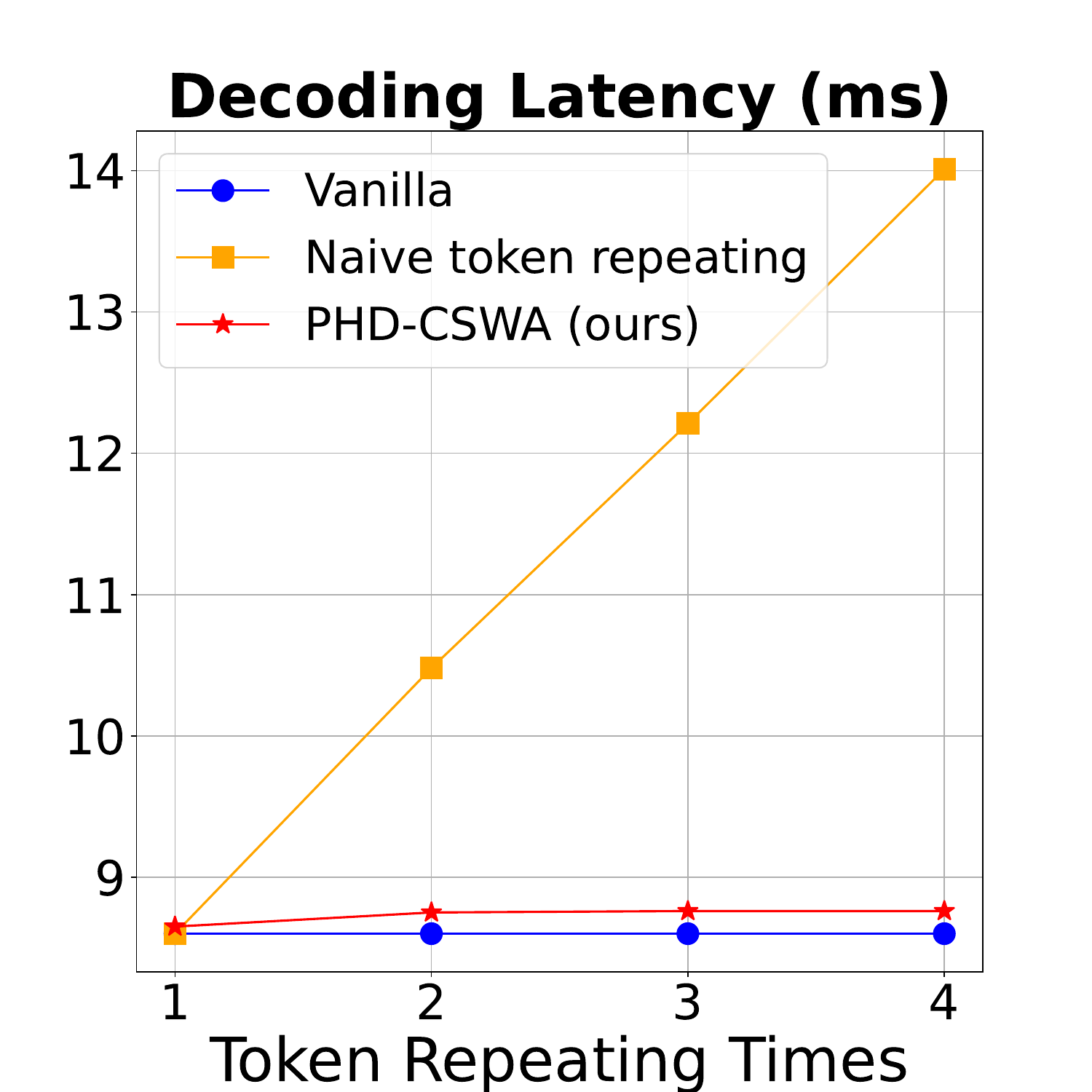}
        \caption{Decoding latency.}
        \label{fig:151m_decode}
    \end{subfigure}
    \caption{The length scaling curve on a 151M sized model. We repeat the training sequence 1/2/3/4 times on the same model architecture and train them for 100B tokens. The training loss and downstream accuracy scale robustly w.r.t. the token repeating times. For repeated training sequence, we only use the final copy of token for next token prediction loss.}
    \label{fig:pretrain_length_scaling}
\end{figure}

More essentially, we present that length scaling can also be achieved in pre-training. Unlike previous works~\cite{mohtashami2023cotformer,tack2025llm}, we simply repeat the input tokens 1/2/3/4 times without post-processing on middle layer hidden states. We observe both the loss scaling and performance scaling trend w.r.t. the token repeating times, which is shown in Figure~\ref{fig:151m_trainloss} and Figure~\ref{fig:151m_hella}. However, naively repeating input tokens incurs significant inefficiencies in inference. The key obstacle arises from the linearly increased KV cache size due to token repetition, which introduces both memory pressure from the KV cache footprint, super linear increase in pre-filling time (shown in Figure~\ref{fig:151m_prefill}), and linear increase in decoding latency (shown in Figure~\ref{fig:151m_decode}).

To address these challenges,
We present a novel inference-friendly length scaling approach.
Our key contribution is the Parallel Hidden Decoding Transformer (\textit{PHD}-Transformer), which maintains the same KV cache size as the vanilla transformer while enabling effective length scaling.
The \textit{PHD}-Transformer achieves this through an innovative KV cache management strategy.
Specifically,
denote the first tokens as \textit{original tokens}, and the repeated tokens as \textit{hidden decoding tokens}, 
we exclusively retain the KV cache generated from \textit{original tokens} for long-range dependency modeling and immediately discard the KV cache of \textit{hidden decoding tokens} after their use in next-token prediction.
This approach provides an identical KV cache size to the vanilla Transformer while delivering substantial inference acceleration compared to naive token repeating (shown in Figure~\ref{fig:151m_decode}).

To better preserve the performance benefits from the KV cache of \textit{hidden decoding tokens},
we introduce \textit{PHD-SWA} (Sliding Window Attention).
This variant maintains a local sliding window cache of these tokens,
achieving notable performance improvements while requiring only $\mathcal{O}(1)$ additional KV cache memory.
However, we notice that the KV cache of \textit{hidden decoding tokens} exhibits sequential dependencies in \textit{PHD-SWA},
which leads to a linear increase in the pre-filling time.
To address this,
we propose \textit{PHD-CSWA} (Chunk-wise Sliding Window Attention), which restricts the sequential dependencies within each chunk.
Therefore, \textit{PHD-CSWA} significantly reduces the pre-filling time as only the pre-filling time of the last chunk is linearly increased (shown in Figure~\ref{fig:151m_prefill}).

In summary, we propose our parallel hidden decoding-transformer series, including \textit{PHD}, \textit{PHD-SWA}, and \textit{PHD-CSWA}.
To our knowledge, this work first presents the topic of efficient pre-training length scaling. Our contributions are summarized as follows:
\begin{itemize}
    \item We propose our \textit{PHD}-Transformer series, which not only demonstrates the effectiveness of length scaling but also provides novel solutions to prevent the linear growth of KV cache size. 
    \item Through comprehensive empirical evaluation, we demonstrate that our PHD-Transformer series delivers substantial accuracy improvements across multiple benchmarks while maintaining acceptable computational overhead in both the pre-filling and decoding phases.
\end{itemize}



%

\section{Approach}

\begin{figure}[tp]
    \centering
    \includegraphics[width=0.7\linewidth]{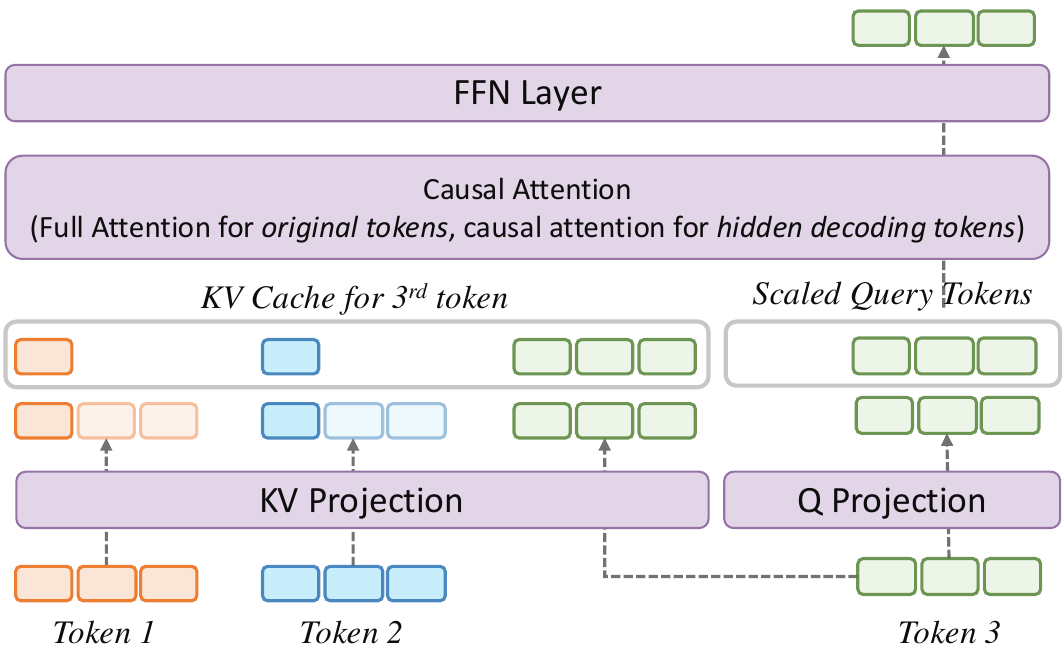}
    \caption{Overview of the transformer block in \textit{PHD}. Specifically, the input tokens are repeated multiple times fed into the transformer block simultaneously. The \textit{original tokens} generate KV cache that can be attended to by all the following tokens, while the \textit{hidden decoding tokens} only generate KV cache that can be attended to within the current tokens (\textit{Token 3} in the Figure). \textbf{We only utilize the final copy of token for next token prediction loss.}}
    \label{figure:overview}
\end{figure}

In this section, we propose our \textit{PHD}-Transformer series, including \textit{PHD}, its sliding window attention variant \textit{PHD-SWA}, and its chunk-wise sliding window attention variant \textit{PHD-CSWA}.

\subsection{Notations}
We set up the notations in this section. Assume a training sample $x \in X$ contains $t$ tokens $\{x_1, x_2, ..., x_t\}$, where the hidden representation of each token can be represented as $\mathbf{h} = \{h_1, h_2, ..., h_t\}$, and the hidden dimension of each token is $d$.  Let $M_{mn}$ be the attention mask between $x_m$ and $x_n$, the original self-attention within sample $x$ can be written as follows:

\begin{equation}
    \text{Self-Attention}(\mathbf{h}) = \text{softmax}\left(\frac{(W_q \mathbf{h}) (W_k \mathbf{h})^\top \otimes M}{\sqrt{d}}\right) (W_v \mathbf{h}).
\end{equation}

In our proposed \ours, we repeat tokens $K$ times, where we extend the sequence to $$\{x_1^1, x_1^2, ...x_1^K;\ \ x_2^1, x_2^2,...,x_2^K;\ \ ...;\ \ x_t^1,x_t^2, ..., x_t^K\}.$$ For better illustration, we name $x_j^1, 1\leq j\leq t$ as the \textit{origin tokens}, and $x_j^s, 1\leq j \leq t, 2 \leq s \leq K$ as the \textit{hidden decoding tokens}. We denote $M_{mn}^{ij}$ as the attention mask between $x_m^{i}$ and $x_n^{j}$.

\subsection{\ours}
The architecture of \ours\ is mainly presented in Figure~\ref{figure:overview}. Compared to the original transformer, \ours\ keeps the same model architecture and only differs in the input sequence and the design of the attention matrix. Specifically, We only allow the \textit{origin tokens} $x_j^1, 1\leq j\leq t$ to generate the KV cache and can be globally attended to by all tokens, while the KV cache of hidden states is dropped instantly after parallel hidden decoding. The attention matrix strategy is formulated as follows:

\begin{equation}
    \text{$M_{mn}^{ij}$} = 
    \begin{cases} 
    1, & \textit{if $i = 1$ and $m < n$} \\
    1, & \textit{if $i \leq j$ and $m = n$} \\
    0. & \textit{otherwise} 
    \end{cases}
    \label{eq:attn_mask}
\end{equation}

Our design achieves identical KV cache size and memory access patterns as the original Transformer during inference.
While requiring $K$ times FLOPs, these computations can be processed in parallel, resulting in minimal latency overhead in memory-bound inference scenarios.
The key advantage of architecture stems from the decoupling between \textit{original tokens} and \textit{hidden decoding tokens}.
During pre-filling, only \textit{original tokens} require computation.
This design ensures pre-filling time is the same as the vanilla transformer and remains constant regardless of the scaling factor $K$.
For loss calculation, we simply use the final copy of tokens for next token prediction. In summary, we use \textit{the first copy of token for KV cache generation}, while use \textit{the final copy of token for next token prediction}.

\subsection{Kernel Design}
\begin{figure}
    \centering
    \includegraphics[width=0.9\linewidth]{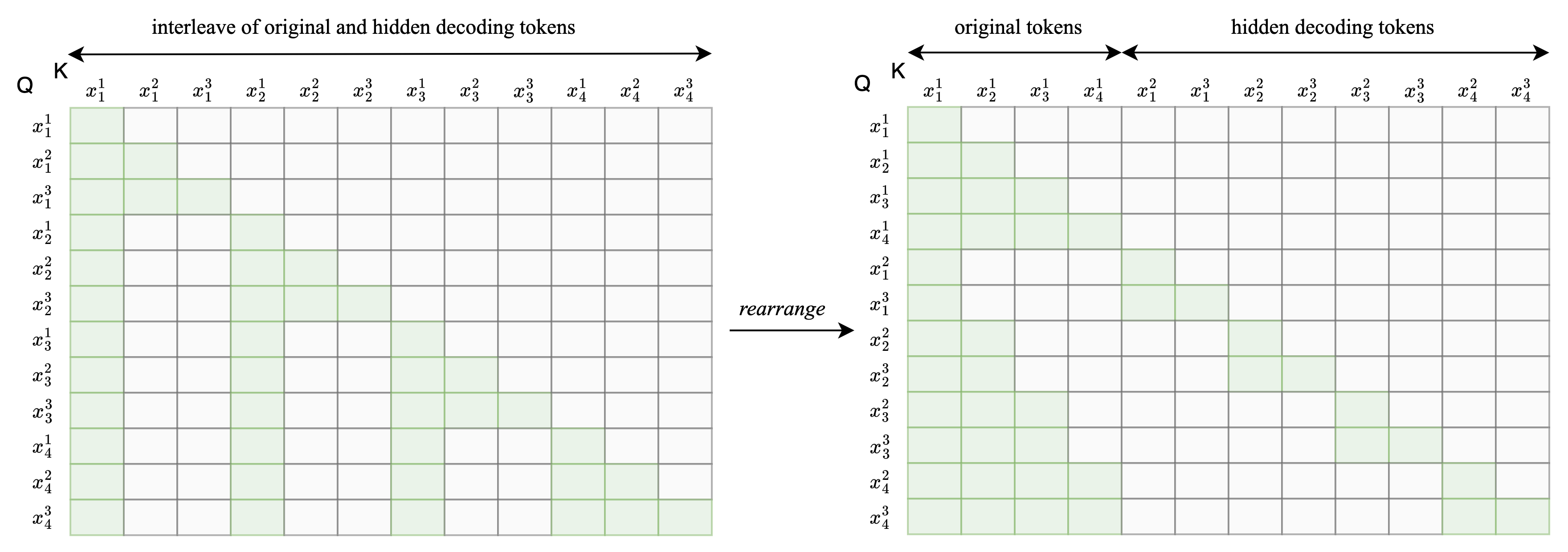}
    \caption{The attention matrix in \ours. The interleaving of \textit{original tokens} and \textit{hidden decoding tokens} introduce very sparse attention matrix that is not device friendly. We rearrange the input sequence and split the original tokens and hidden decoding tokens into two groups. In this way, we group the un-attended attention positions in a continuous block, which is efficient for optimization.}
    \label{fig:naive_attn}
\end{figure}

Naive implementation of $M_{mn}^{ij}$ results into $K^2$ times computation increase in the attention layers and $K$ times increase in the FFN layers. However, since the attention is sparsely computed, the $\mathcal{O}(K^2)$ attention can be largely reduced. Consequently, we split the \textit{original tokens} and \textit{hidden decoding tokens} into two groups, and concatenate them together. Figure~\ref{fig:naive_attn} shows an example of $K=3$, where we can get one sequence of $t$ original tokens and one sequence of $2t$ hidden decoding sequence. By rearranging the token positions, we keep the attention positions that are masked in one continuous block, which leads to optimization in attention computation, reducing the complexity of attention computation to $\mathcal{O}(K)$.


\subsection{\oursswa\ and \ourscswa}
\begin{figure}
    \centering
    \includegraphics[width=0.9\linewidth]{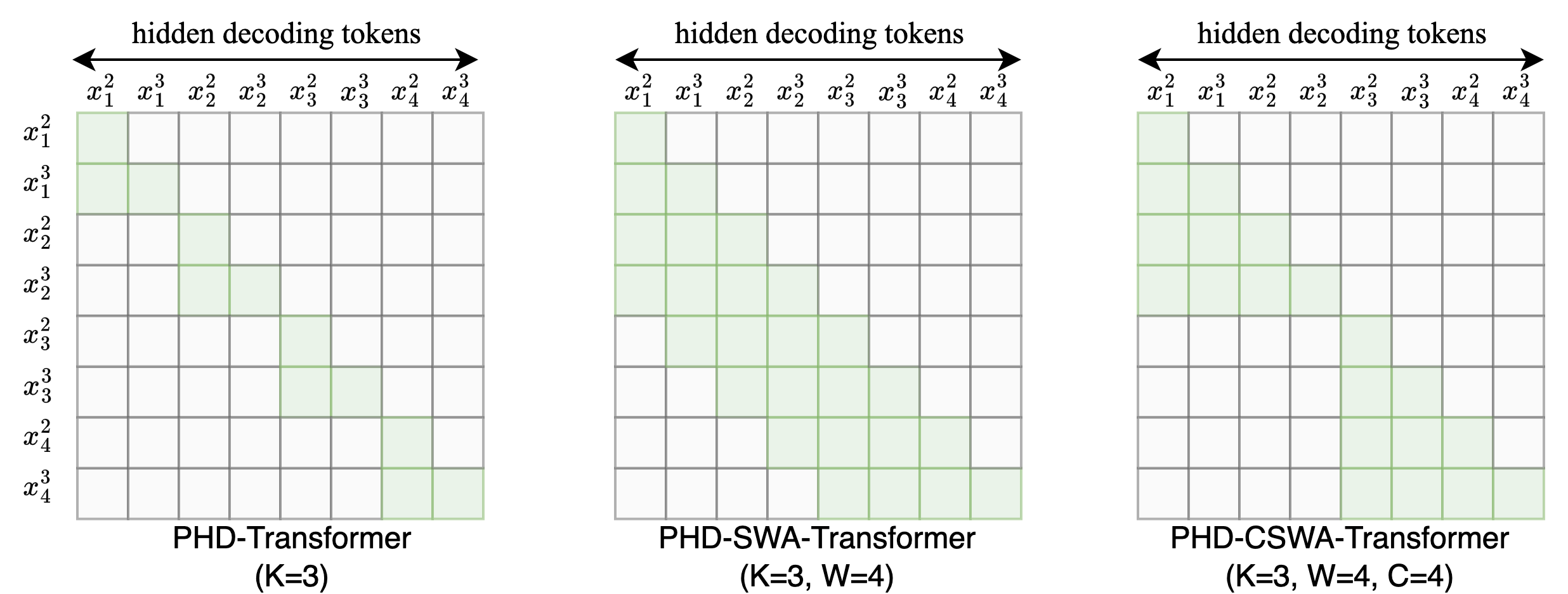}
    \caption{Comparison of the attention matrices in \textit{PHD}, \textit{PHD-SWA} and \textit{PHD-CSWA}. In this figure, we set the repeating times $K$ to 3, which means there are 2 hidden decoding tokens in each attention matrix, and set the window size $W$ to 4 and chunk size $C$ to 4.}
    \label{fig:attention_compare}
\end{figure}

Compared to naive token repeating,
our \textit{PHD}-Transformer achieves length scaling while maintaining the original KV cache size.
However, we empirically observe that preserving some KV cache for \textit{hidden decoding tokens} yields significant performance benefits.
To capture these benefits while maintaining efficiency, we introduce PHD-SWA, which implements sliding window attention restricted to $W$ preceding \textit{hidden decoding tokens}.
As illustrated in Figure~\ref{fig:attention_compare},
the attention pattern combines global access to original tokens with local access to the $W$ most recent \textit{hidden decoding tokens}.
This modified attention mechanism achieves notable performance improvements while only requiring $\mathcal{O}(1)$ additional KV cache memory.

While the sliding window approach in \textit{PHD-SWA} enhances model performance,
it incurs a $K$ times pre-filling overhead caused by sequential dependencies in the KV cache of \textit{hidden decoding tokens}.
To address this, we introduce \ourscswa\ which processes attention within independent chunks.
As demonstrated in Figure~\ref{fig:attention_compare},
\ourscswa\ constrains the sliding window attention to operate within individual chunks.
This architectural innovation reduces the extra pre-filling overhead to just $K$ repetitions within the final chunk, rather than across the entire sequence,
making the additional computational cost practically negligible while preserving the benefits of local attention patterns.

\section{Experiments}

In this section, we present a detailed experimental analysis of our proposed \textit{PHD}. We use OLMo2~\cite{olmo20242} as the codebase of all our experiments. The hyperparameters and details are illustrated in each subsection correspondingly.  All model variants are named in the format of \textit{model-type-K-W-C} format for better illustration.

\subsection{Evaluation Metrics}

To evaluate the performance of our proposed \textit{PHD-Transformer} series, we evaluate them on the following set of open benchmarks, including ARC~\cite{clark2018arc}, HellaSwag~\cite{zellers2019hellaswag}, PIQA~\cite{bisk2020piqa}, Winogrande~\cite{sakaguchi2021winogrande}, MMLU~\cite{hendrycks2020measuring},
and CommonsenseQA~\cite{talmor2018commonsenseqa}.

\subsection{Main Results}

\begin{figure}
    \centering
    \includegraphics[width=0.9\linewidth]{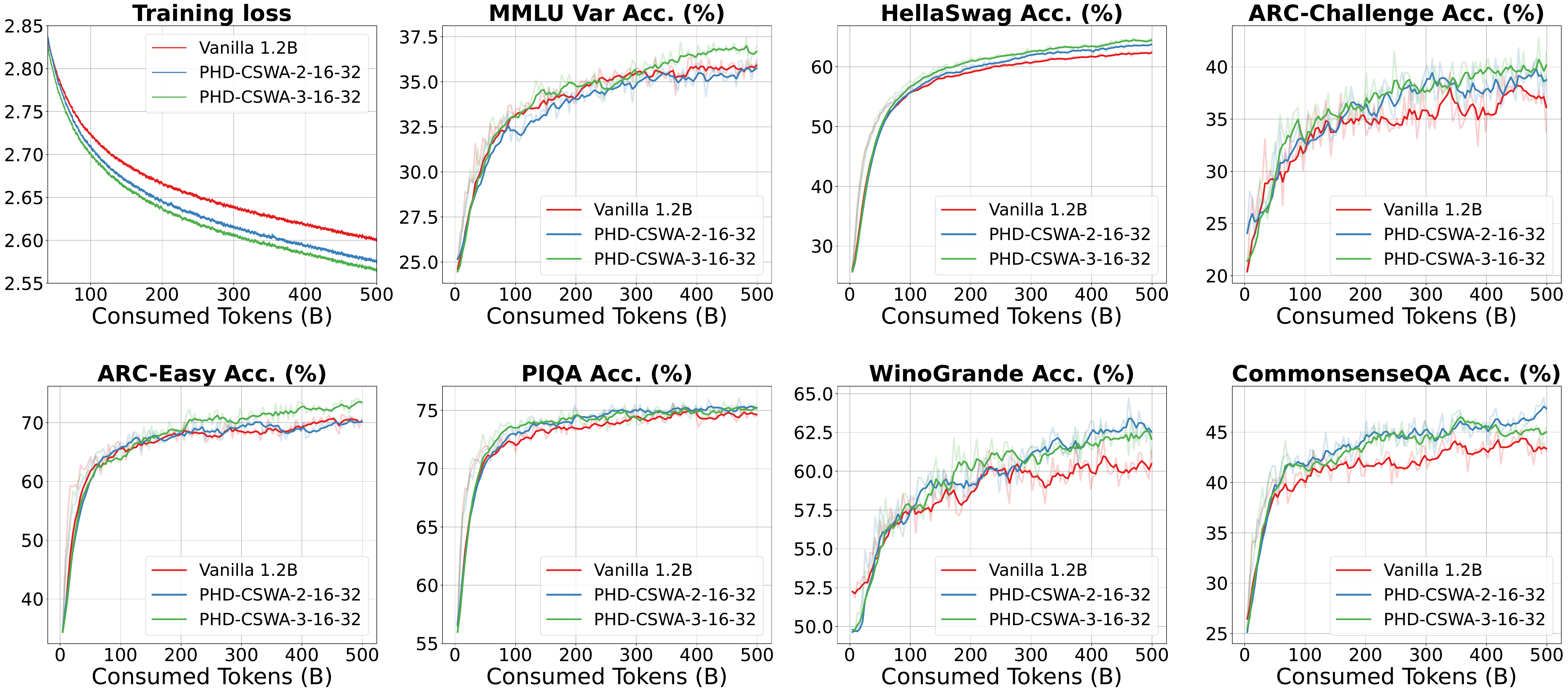}
    \caption{Training curves of \ourscswa variants and baseline model on OLMo2-1.2B. We smooth these metrics via exponential moving average with weight 0.99 for loss and 0.7 for downstream tasks.}
    \label{fig:main_1b2}
\end{figure}

In this section, we present the main results to show the efficacy of our \ourscswa\ as it is the most practical and effective strategy, which introduces steady performance improvement with acceptable overhead.

\paragraph{Training Details} We use the 1.2B-sized model, which is a 16-layer dense model. The hidden dimensions of each token is set to 2048, and the hidden size of the FFN layer is set to 16384. We use the Group-Query Attention (GQA)~\cite{ainslie2023gqa}, which contains 32 query heads and 8 key/value heads, where the hidden dimension of each head is set to 64. We train this model for 500B tokens.
For the settings of our proposed \ours\ series, we pretrain two-variants of \ourscswa\ listed as follows:
\begin{itemize}
    \item \textit{PHD-CSWA-2-16-32}, where we repeat the training tokens twice. We keep a local window of 16 tokens and set the chunksize to 32 tokens.
    \item \textit{PHD-CSWA-3-16-32}, where we repeat the training tokens 3 times. The local window size and chunk size are set to the same values with \textit{PHD-CSWA-2-16-32}.
\end{itemize}

\paragraph{\ourscswa\ introduce consistent performance improvement across various benchmarks.} We presents the training curves in Figure~\ref{fig:main_1b2} and the main results in Table~\ref{tab:main_1b2}.
Our proposed \textit{PHD-CSWA-2-16-32} introduces an average of 1.5\% accuracy improvement across these benchmarks and 0.025 decrease in training loss, while \textit{PHD-CSWA-3-16-32} introduces an average of 2.0\% accuracy improvement and 0.034 decrease in training loss. 

\subsection{Ablation Studies}
We perform comprehensive ablation studies using a 550M-sized dense transformer model. The architecture consists of 16 layers with the hidden dimension set to 1536, and the hidden size of the FFN layer set to 8192.
We adopt GQA, with 16 query heads and 4 key/value heads.
We train each model for 300B tokens.

\begin{table}[tp]
    \centering
    \small
    \setlength{\tabcolsep}{4pt}
    \begin{tabular}{c|c|ccccccc|c}
    \toprule
    & loss$\downarrow$ & MMLU-V$\uparrow$ & Hella.$\uparrow$ & ARC-C$\uparrow$ & ARC-E$\uparrow$ & PIQA$\uparrow$ & Wino.$\uparrow$ & Comm.$\uparrow$ & Avg.$\uparrow$ \\
    \midrule
        Vanilla 1.2B & 2.601 & 35.9 & 62.3 & 36.1 & 70.3 & 74.6 & 60.5 & 43.4 & 54.7 \\
        \textit{PHD-CSWA-2-16-32} 1.2B& 2.576 & 35.8 & 63.7 & 38.8 & 70.2 & 75.2 & 62.5 & 47.4 & 56.2 \\
        \textit{PHD-CSWA-3-16-32} 1.2B& 2.567 & 36.7& 64.5 & 40.2 & 73.5 & 75.2 & 62.1 & 45.0 & 56.7 \\
    \bottomrule
    \end{tabular}
    \caption{
    Performance evaluation of 1.2B parameter dense models using our PHD-CSWA variants with scaling factors $K\in \{2, 3\}$.
    The window size $W$ is set to 16 and the chunk size $C$ is set to 32 for our proposed \ourscswa\ variants.
    Evaluated benchmarks include: MMLU Var (MMLU-V), Hellaswag (Hella.), ARC-Challenge (ARC-C), ARC-Easy (ARC-E), PIQA, Winogrande (Wino.), and Commonsense QA (Comm.).
    }
    \label{tab:main_1b2}
\end{table}

\subsubsection{Chunk-wise Sliding Window Attention}
Since the chunk-wise sliding window attention is a compensation for accelerating the pre-filling stage, we are interested in the performance drop that CSWA may introduce.
Meanwhile, we also study the performance difference when different window sizes and chunk sizes are chosen.

\paragraph{Training Details.} We keep the scaling factor $K=2$,
and vary the window size in $\{1, 2, 4, 16\}$.
For the largest window size ($W=16$), we conduct additional experiments to study the effect of different chunk sizes, where we set the chunk size to 16, 32, and no chunk at all.


\begin{figure}[h]
    \centering
    \begin{minipage}[b]{0.48\textwidth}
        \begin{subfigure}[b]{0.49\textwidth}
            \includegraphics[width=\linewidth]{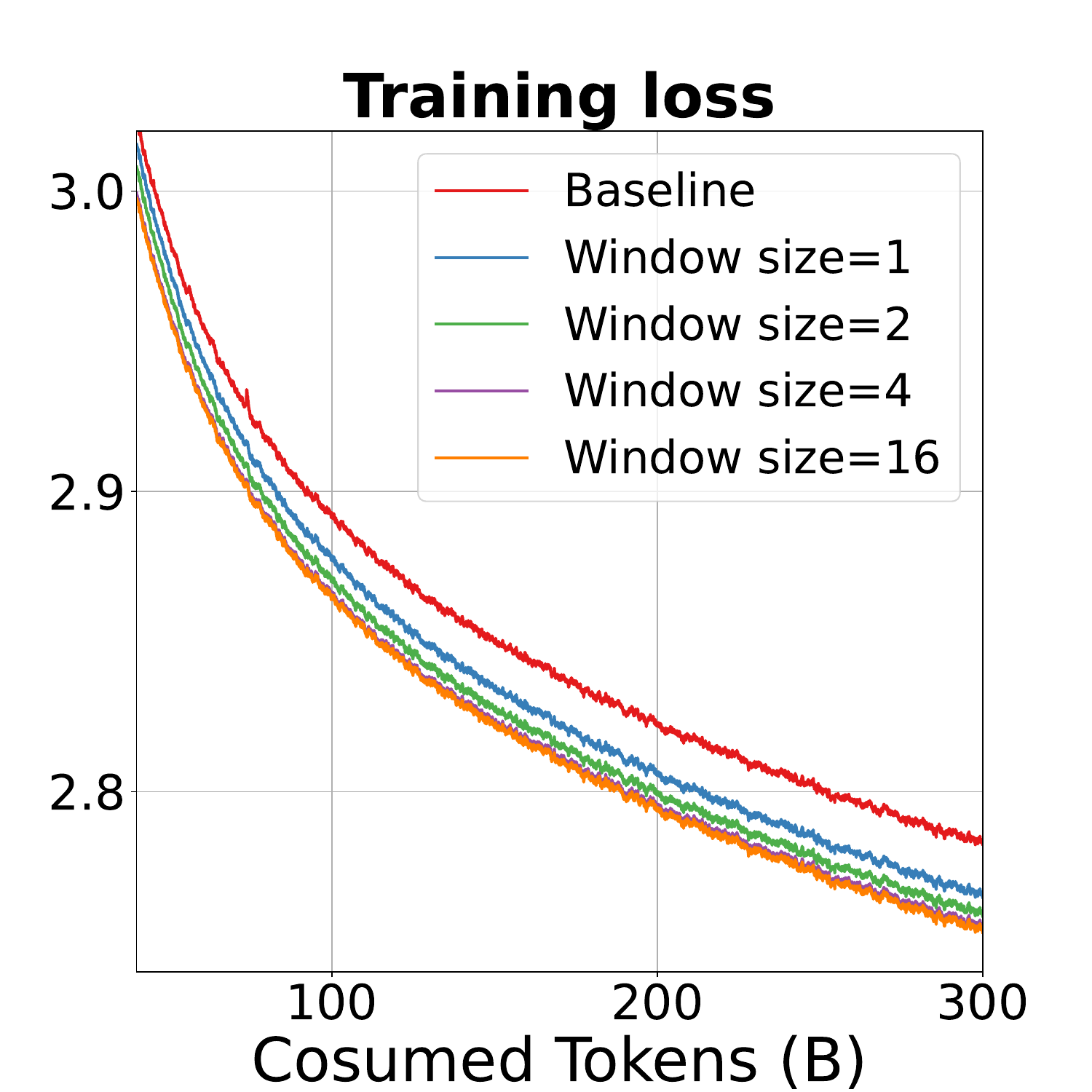}
            \caption{}
            \label{fig:window_ablation_trainingloss}    
        \end{subfigure}
        \begin{subfigure}[b]{0.49\textwidth}
            \includegraphics[width=\linewidth]{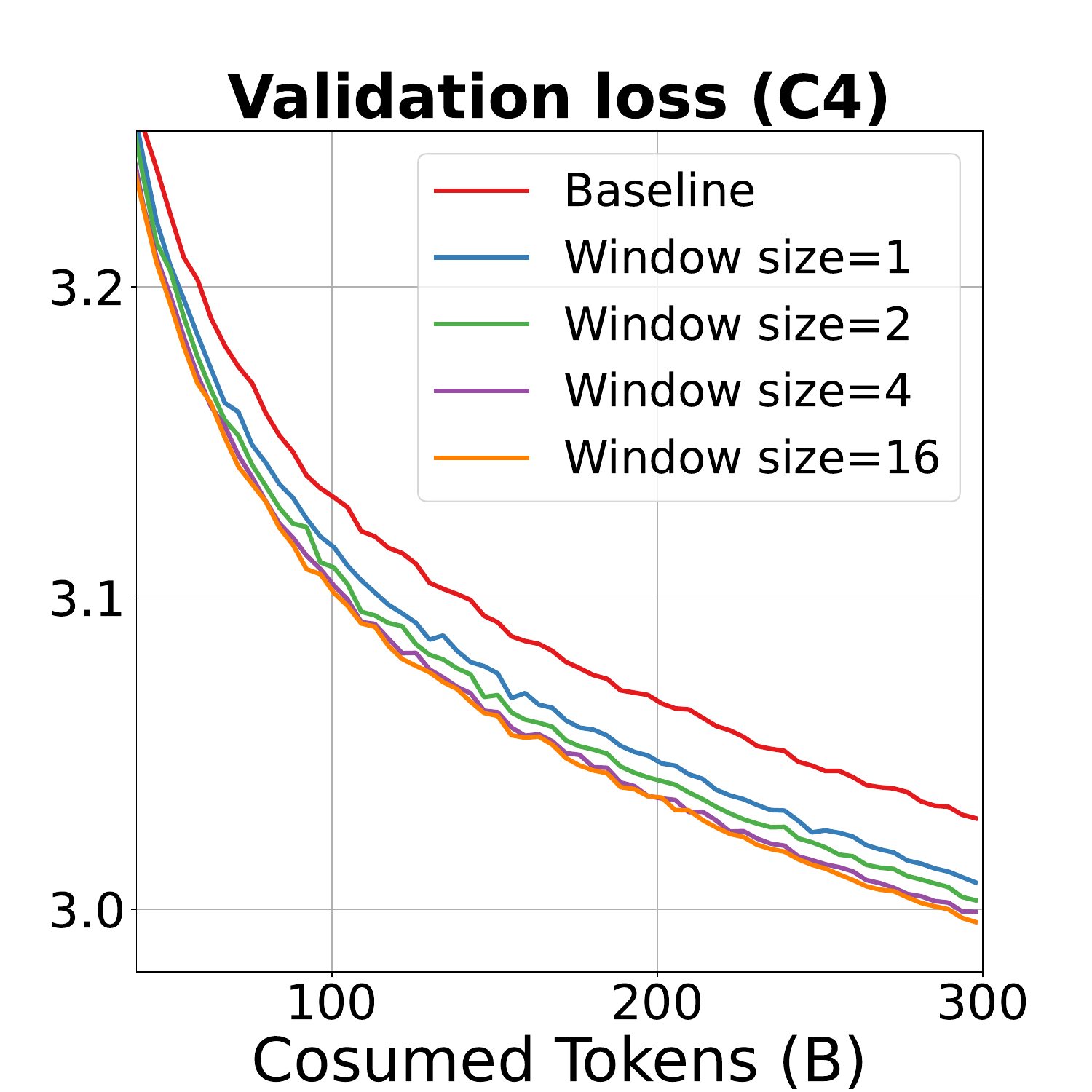}
            \caption{}
            \label{fig:window_ablation_valloss}    
        \end{subfigure}
        \caption{Ablation studies on window size.}
        \label{fig:window_ablation}
    \end{minipage}
    \hspace{0.01\textwidth} 
    \begin{minipage}[b]{0.48\textwidth}
        \begin{subfigure}[b]{0.49\textwidth}
            \includegraphics[width=\linewidth]{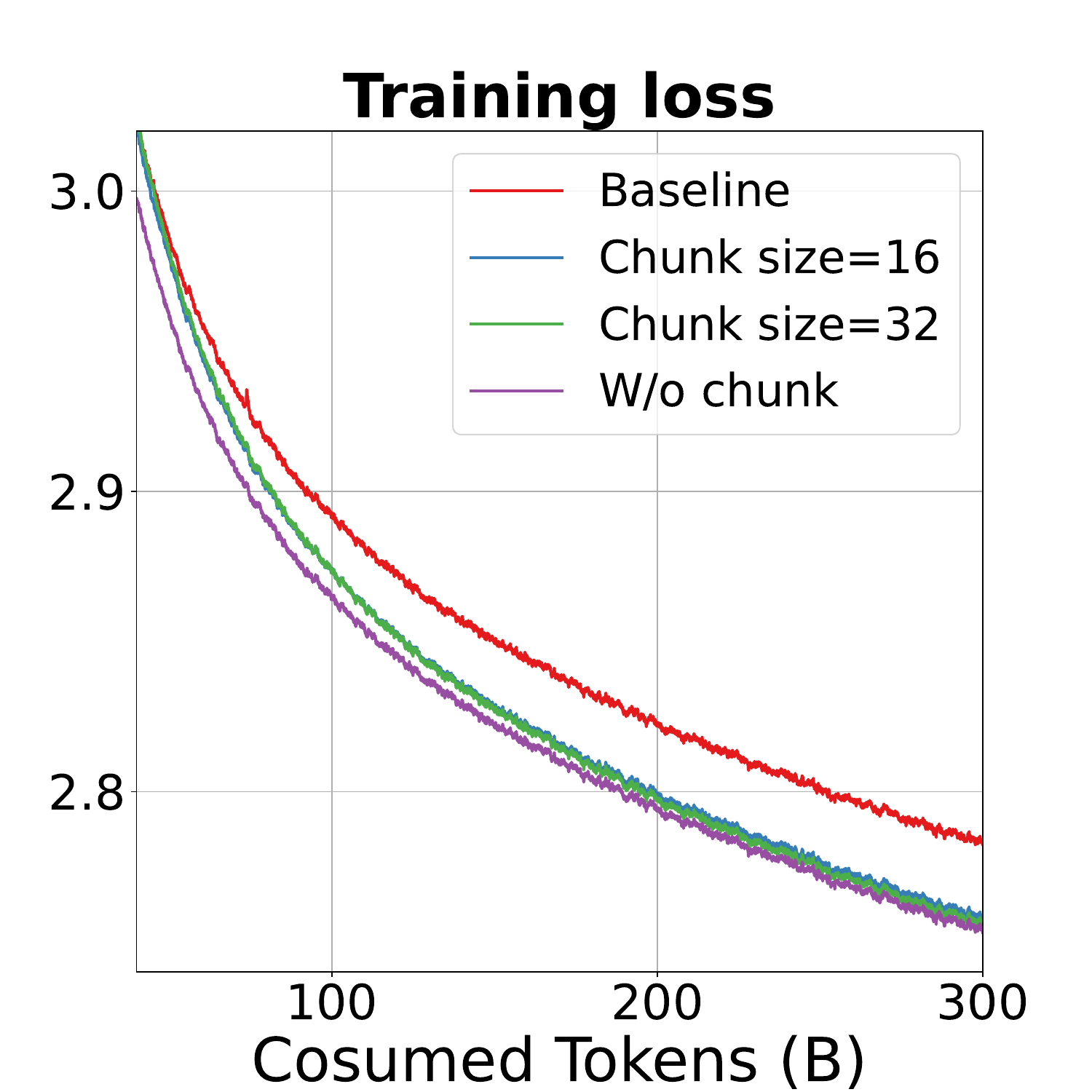}
            \caption{}
            \label{fig:chunk_ablation_trainingloss}
        \end{subfigure}
        \begin{subfigure}[b]{0.49\textwidth}
            \includegraphics[width=\linewidth]{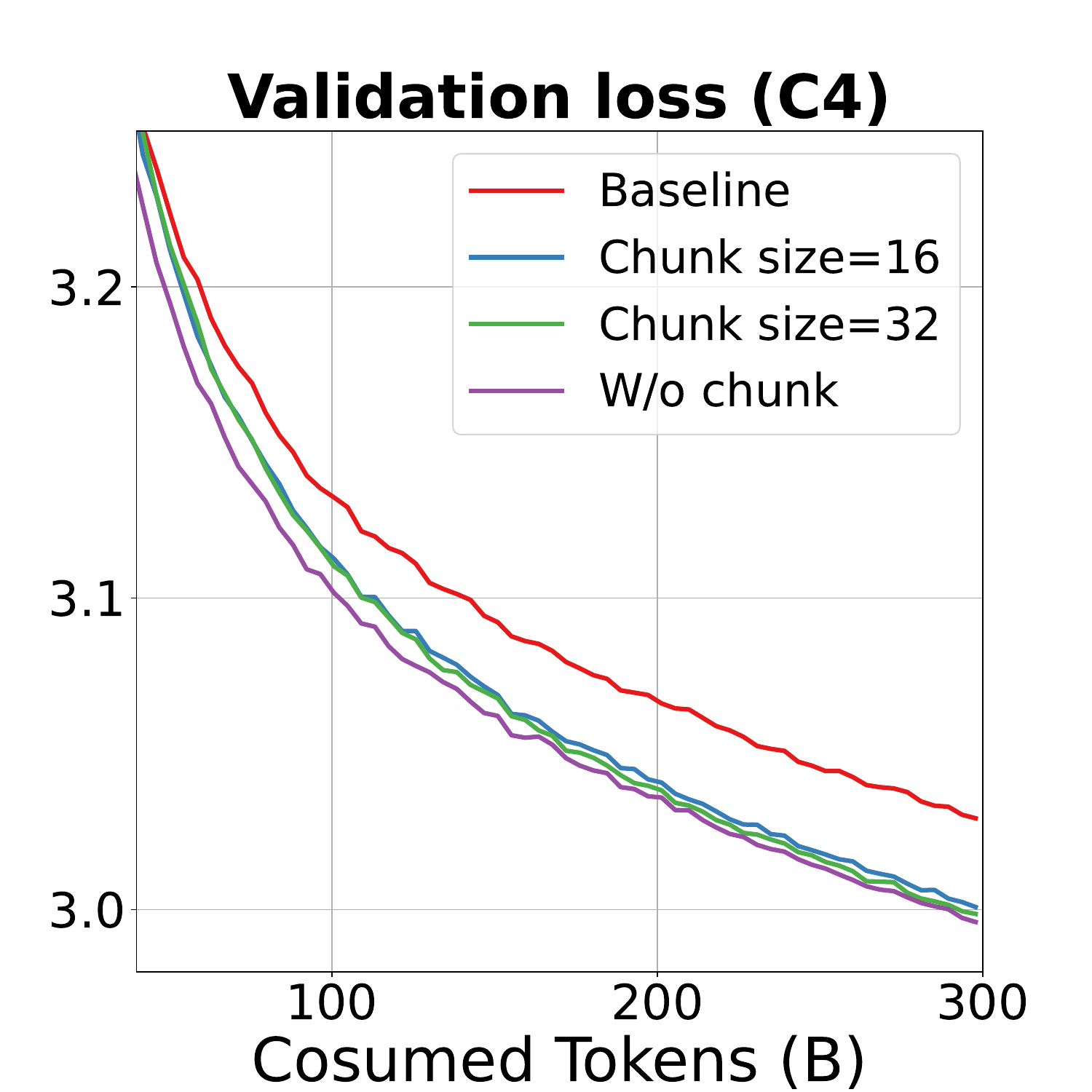}
            \caption{}
            \label{fig:chunk_ablation_valloss}
        \end{subfigure}
        \caption{Ablation studies on chunk size.}
        \label{fig:chunk_ablation}
    \end{minipage}
\end{figure}
 

%
\paragraph{Large window size leads to better performance, but the improvement converges very fast.}
As shown in Figure~\ref{fig:window_ablation},
While expanding the window from $W=1$ to $W=4$ yields significant reductions in both training and validation loss, further increasing to $W=16$ provides only marginal improvements.
This suggests that maintaining a small window of local KV cache for \textit{hidden decoding tokens} achieves nearly optimal performance while remaining computationally efficient and hardware-friendly.

\paragraph{Introducing chunks for reducing pre-filling overhead leads to little performance degradation.}
As shown in Figure~\ref{fig:chunk_ablation}, the introduction of chunks causes only negligible differences in both training and validation losses.
We also observe a consistent trend where larger chunk sizes yield progressively better results. Based on this analysis, we select $C=32$ as the optimal balance between computational efficiency and model performance for all subsequent experiments.

\subsection{Decoding Token Scaling}
In this section, we analyze the scaling performance of \ours\ and \oursswa\ to analyze the performance of scaling decoding computation.

\paragraph{Training Details}
We use the same 550M model configuration in this section. We set the window size $W$ to 16 and vary the scaling factor $K$ in $\{2, 3, 5\}$. For local window size, we set the window size to 16 for all experiments.

\begin{figure}[h]
    \centering
    \begin{subfigure}[b]{0.23\textwidth}
        \includegraphics[width=\linewidth]{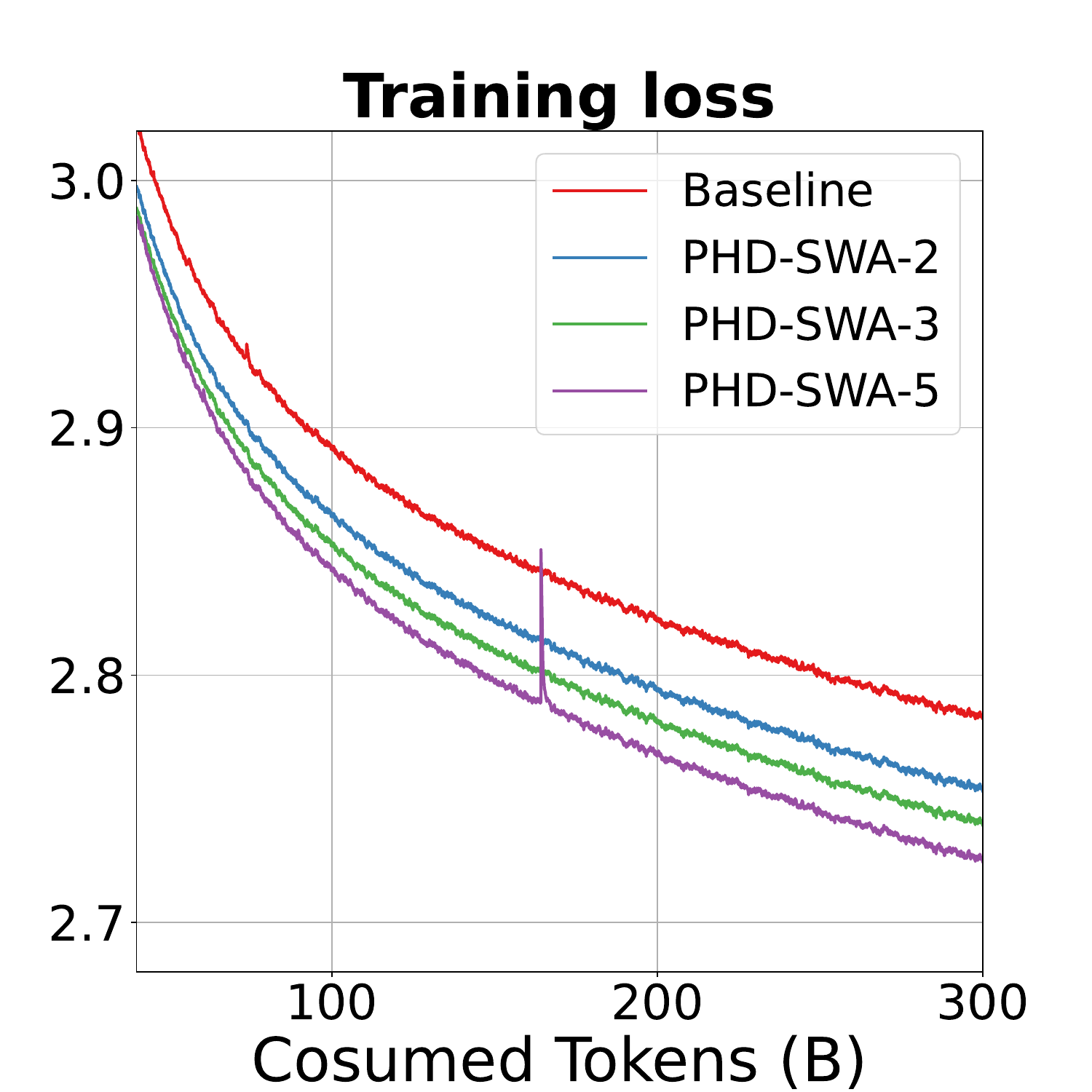}
        \caption{Training loss.}
        \label{fig:dec_token_ablation_withwindow_trainingloss}
    \end{subfigure}
    \hspace{0.01\textwidth} 
    \begin{subfigure}[b]{0.23\textwidth}
        \includegraphics[width=\linewidth]{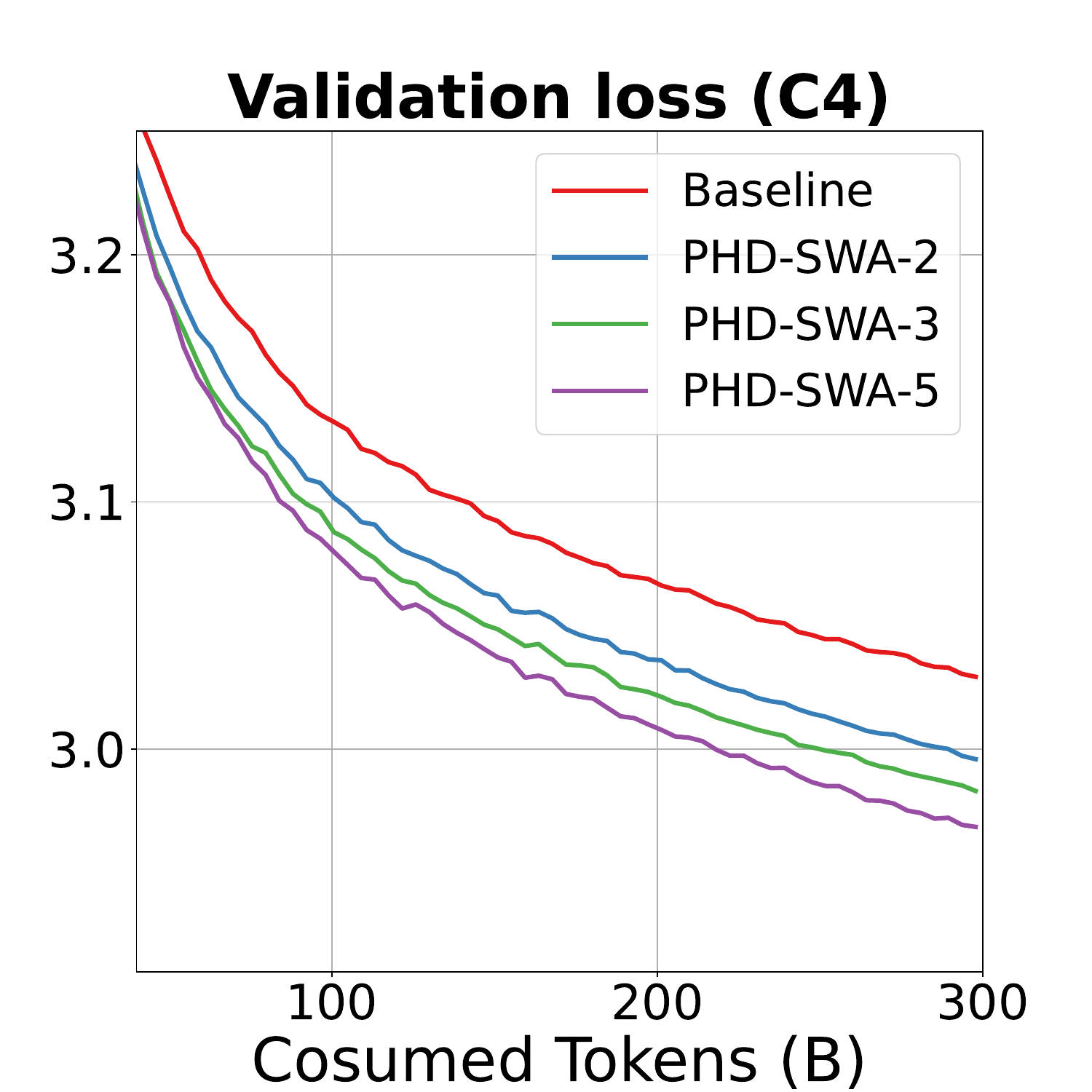}
        \caption{Validation loss.}
        \label{fig:dec_token_ablation_withwindow_valloss}
    \end{subfigure}
    \hspace{0.01\textwidth} 
    \begin{subfigure}[b]{0.23\textwidth}
        \includegraphics[width=\linewidth]{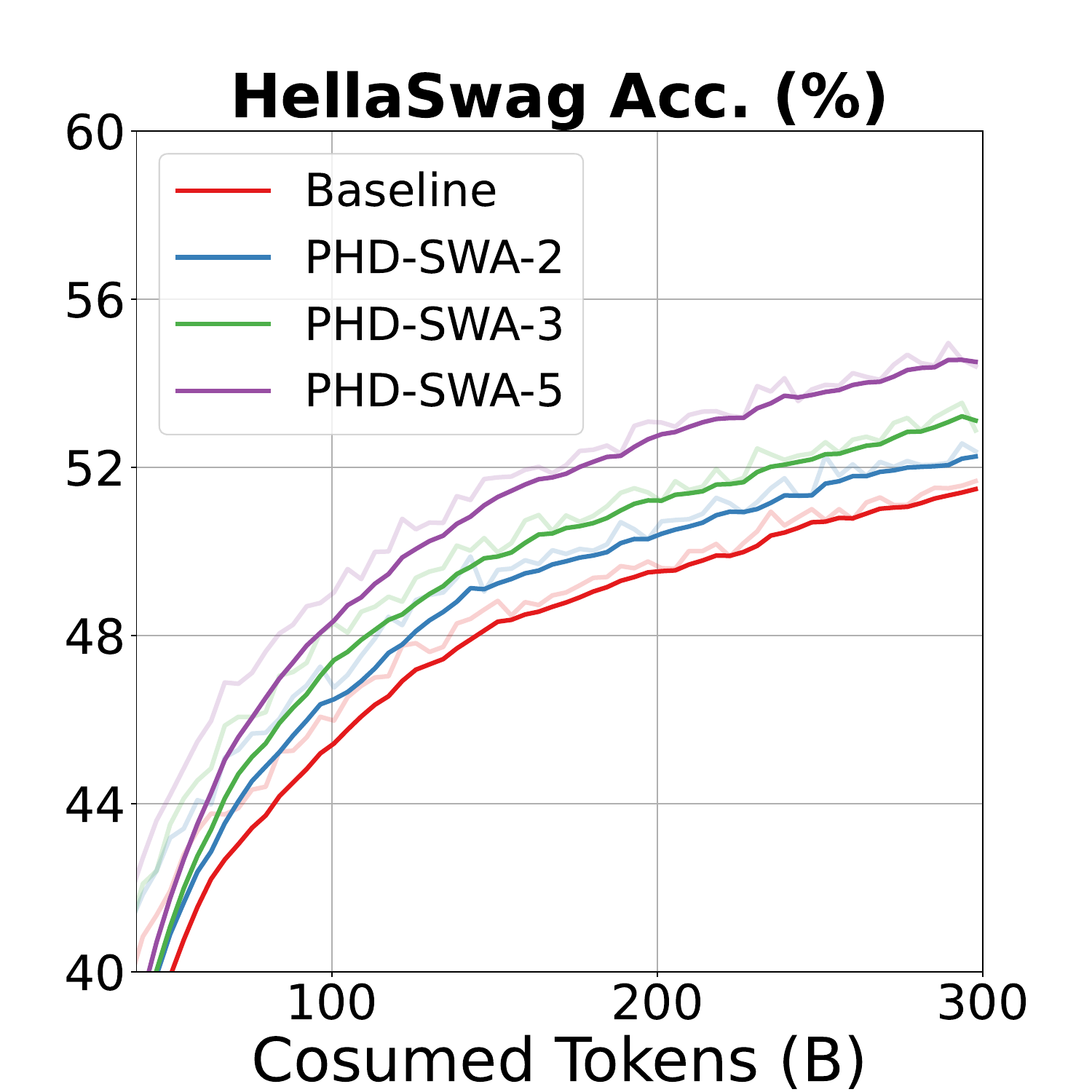}
        \caption{Hellaswag accuracy.}
        \label{fig:dec_token_ablation_withwindow_hellaswag}
    \end{subfigure}
    \hspace{0.01\textwidth} 
    \begin{subfigure}[b]{0.23\textwidth}
        \includegraphics[width=\linewidth]{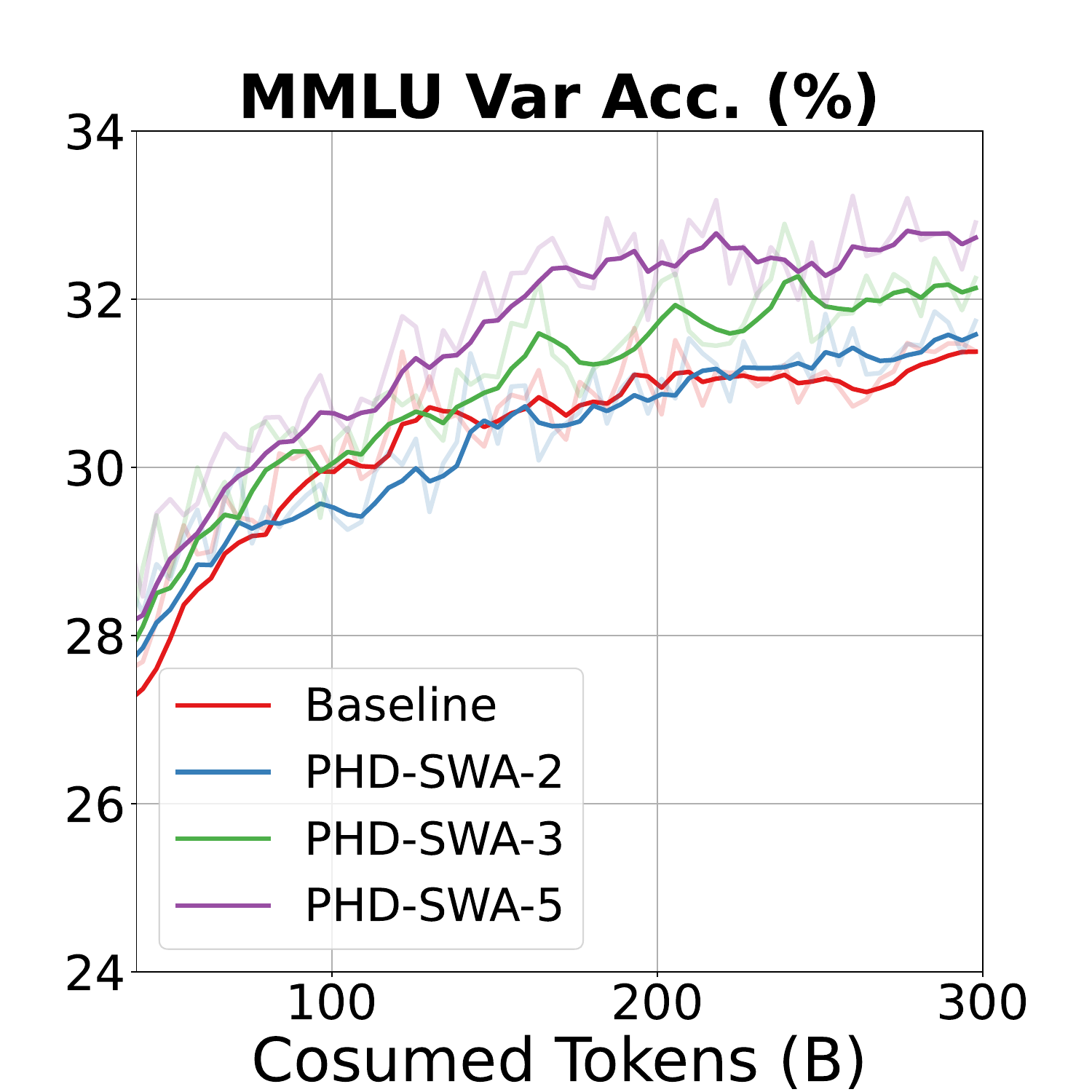}
        \caption{MMLU-V accuracy.}
        \label{fig:dec_token_ablation_withwindow_mmlu}
    \end{subfigure}
    \caption{The scaling behavior of \textit{PHD-SWA-{$K$}-{16}-{$\infty$}}.}
    \label{fig:abl_window_chunk}
\end{figure}

\paragraph{Performance of \oursswa\ scales effectively when increasing the scaling factor.} As shown in Figure~\ref{fig:abl_window_chunk}, we can see that using a fixed window size, the loss curve and downstream performance scales effectively w.r.t. the token repeating times. By setting the scaling factor to 5, we achieve near 0.06 loss decrease with notable downstream performance improvement.
Quantitative results in Table~\ref{tab:scale_decode_token} reveal an average accuracy improvement of 1.8\% across all benchmarks when scaling to $K=5$, confirming the effectiveness of our approach for more aggressive scaling.

\begin{table}[tp]
    \centering
    \small
    \setlength{\tabcolsep}{4pt}
    \begin{tabular}{c|c|ccccccc|c}
    \toprule
    & Loss$\downarrow$ & MMLU$\uparrow$ & Hella.$\uparrow$ & ARC-C$\uparrow$ & ARC-E$\uparrow$ & PIQA$\uparrow$ & Wino.$\uparrow$ & Comm.$\uparrow$ & Avg.$\uparrow$ \\
    \midrule
        Vanilla 550M & 2.782 & 31.4 & 51.5 & 30.1 & 62.5 & 71.9 & 56.8 & 40.6 & 49.3 \\
        \textit{PHD-SWA-2-16-$\infty$} 550M & 2.753 & 31.6 & 52.3 & 31.5 & 64.2 & 71.5 & 56.6 & 42.7 & 50.1 \\
        \textit{PHD-SWA-3-16-$\infty$} 550M & 2.739 & 32.1 & 53.1 & 31.5 & 64.6 & 71.5 & 56.9 & 41.7 & 50.2 \\
        \textit{PHD-SWA-5-16-$\infty$} 550M & 2.725 & 32.7 & 54.5 & 34.7 & 65.0 & 72.3 & 56.2 & 42.6 & 51.1 \\
    \bottomrule
    \end{tabular}
    \caption{Decoding computation scaling trend on \textit{PHD-SWA}. The downstream performance scales w.r.t. the increase of decoding computation.}
    \label{tab:scale_decode_token}
\end{table}

\subsection{Pre-filling Speed and Decoding Speed}

In this section, we evaluate both the pre-filling time and decoding latency, showing that our proposed \ours\ series introduce only marginal latency overhead during the inference stage.
The experiments are conducted on 550M-sized model using a single A100 GPU, and the decoding batch size is set to 1.

\begin{figure}
    \centering
    \begin{subfigure}[b]{0.35\textwidth}
        \includegraphics[width=\linewidth]{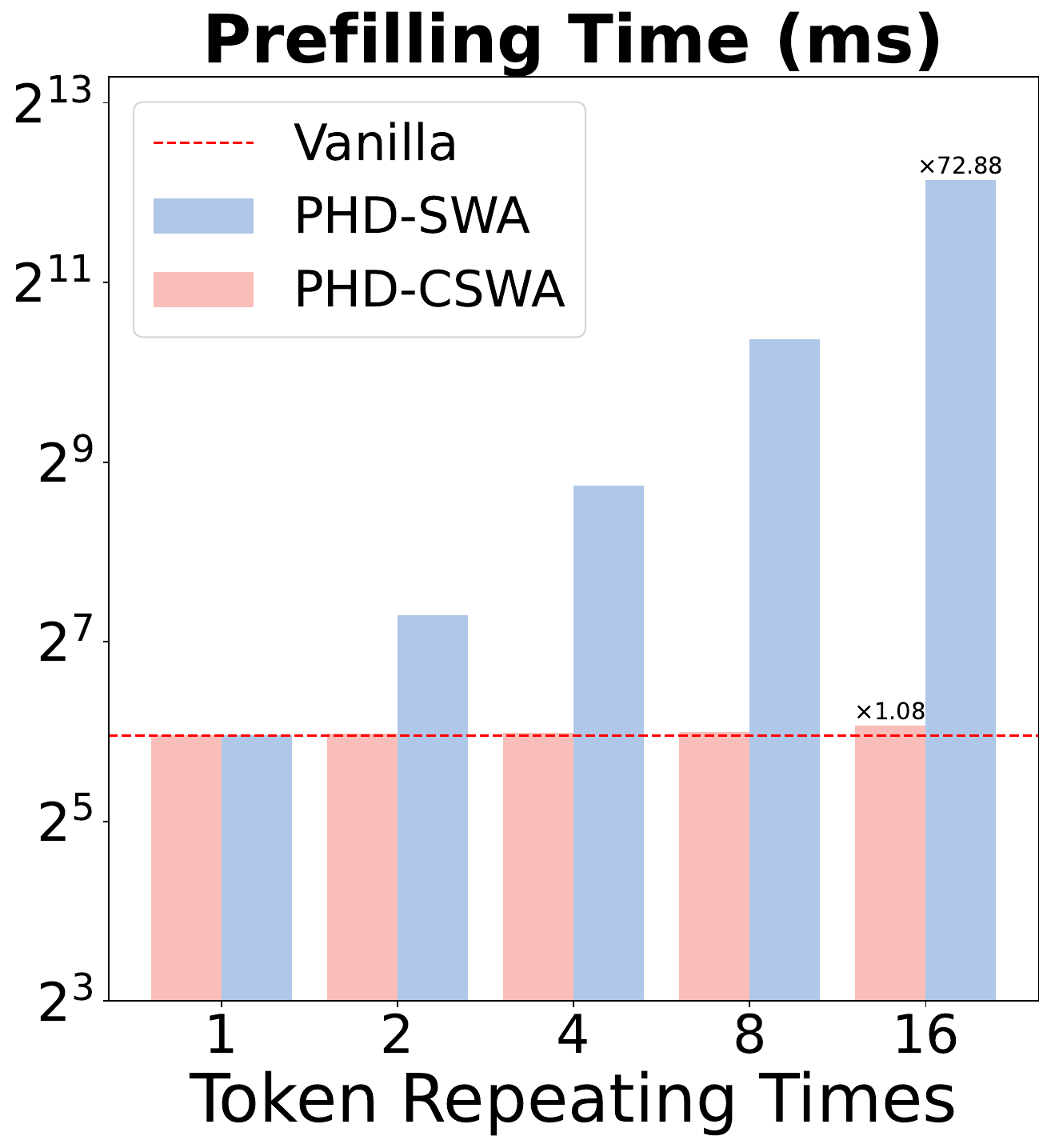}
        \caption{}
        \label{fig:prefill_time}
    \end{subfigure}
    \hspace{0.12\textwidth} 
    \begin{subfigure}[b]{0.35\textwidth}
        \includegraphics[width=\linewidth]{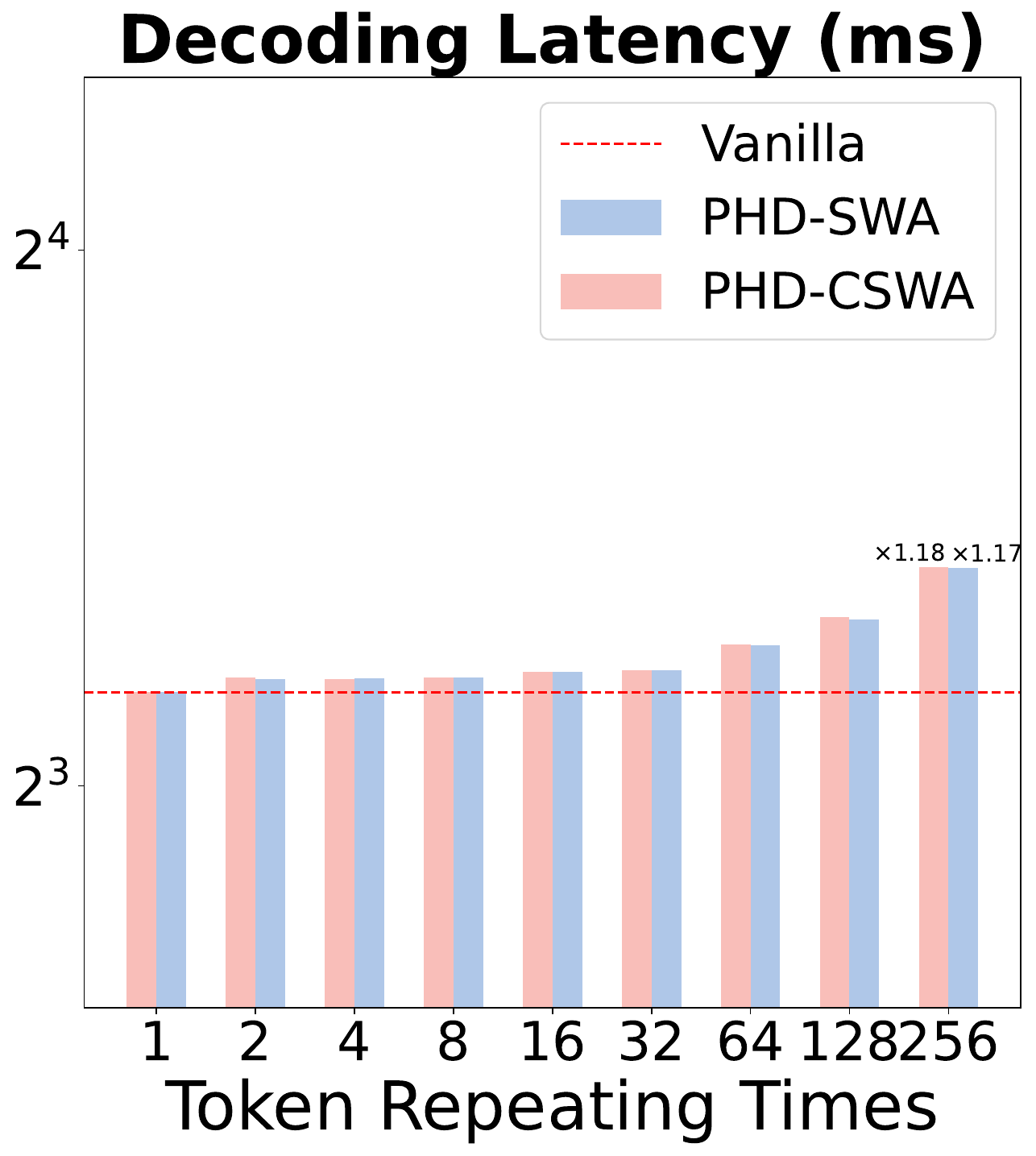}
        \caption{}
        \label{fig:decode_latency}
    \end{subfigure}
    \caption{(a) Prefilling time and (b) decoding latency on different models when the repeating times is varied.}
    \label{fig:cost_time}
\end{figure}

\paragraph{Our proposed \ours\ series introduce marginal inference overhead.} 
As shown in Figure~\ref{fig:prefill_time}, \oursswa\ introduce superlinear pre-filling overhead when the sequence length increases, while \ourscswa\ only incurs a small increase compared to the vanilla baseline.
Figure~\ref{fig:decode_latency} presents the decoding latency when we vary the scaling factor $K$ in $\{1, 2, 4, 8, 16, 32, 64, 128, 256\}$.
Since decoding latency is limited by memory bound,
both \oursswa\ and \ourscswa\ introduces minimal overhead when $K$ is increased.
Specifically,
setting $K$ to 256 only leads to no more than 20\% increase in decoding latency.


\section{Related Works}

The computational and memory challenges of scaling transformer models, especially in terms of attention mechanisms, have led to numerous research efforts focusing on sparse attention patterns, KV cache optimization, and efficient inference techniques. Our Parallel Hidden Decoding Transformer (PHD) builds upon and extends these approaches in novel ways. Below, we categorize and analyze relevant works in this space, drawing comparisons with our approach.

\subsection{Sparse Attention Mechanisms}

Sparse attention techniques aim to reduce the quadratic complexity of attention by focusing on the most informative token interactions. These approaches can be broadly categorized into three types, each with distinct relationships to our work:

\paragraph{Fixed Pattern Sparse Attention.} These methods use predefined patterns to constrain attention computation. Notable examples include sliding window attention in Mistral \citep{jiang2023mistral7b} and Phi-3 \citep{Abdin2024phi3}, which restrict attention to local contexts; dilated attention \citep{child2019sparsetransformers,shi2021sparsebert,ding2023longnet}, which attends to tokens at increasing intervals; and mixed patterns as seen in Longformer \citep{beltagy2020longformer} and BigBird \citep{zaheer2020bigbird}, which combine local attention with global tokens. While our \ours shares the use of predefined attention patterns, particularly in our \oursswa variant which employs sliding window attention for \textit{hidden decoding tokens}, we uniquely apply these patterns only to the repeated tokens, maintaining full attention for the original tokens and thus preserving model quality while gaining efficiency.

\paragraph{Data-Dependent Sparse Attention.} Unlike fixed patterns, these approaches dynamically determine attention patterns based on input characteristics. Quest \citep{tang2024quest} proposes data-dependent block-wise sparse attention that adaptively selects blocks based on token-block similarity. Other examples include SpAtten \citep{wang2021spatten} and SparQ \citep{ribar2023sparq}, which leverage the dynamic nature of attention to predict sparse patterns. While these methods offer adaptive flexibility, they introduce substantial overhead in estimating the sparse patterns, limiting their effectiveness for long-context scenarios. In contrast, our PHD approach uses static patterns that require no runtime estimation overhead, making it more suitable for practical deployment while still achieving performance improvements through increased computational depth.

\paragraph{Training-Native Sparse Attention.} To address the inconsistency between training and inference in post-hoc sparse attention, Native Sparse Attention (NSA) \citep{yuan2025native} incorporates block sparse patterns during the training stage itself. This approach achieves improvements in both downstream performance and efficiency. Similar to NSA, our PHD method integrates efficient attention patterns during training, but unlike NSA's focus on general block sparsity, we specifically target \textit{hidden decoding tokens} with tailored attention patterns, allowing us to maintain the same KV cache footprint as the original transformer while significantly improving performance.

\subsection{KV Cache Optimization}

As models scale to handle longer contexts, KV cache management becomes increasingly critical for efficient inference:

\paragraph{KV Cache Reduction.} Various approaches attempt to reduce KV cache size, including H2O \citep{zhang2023h2o}, which identifies and discards less important KV cache entries; StreamingLLM \citep{xiao2024streamingllm}, which employs sink attention for handling streaming inputs; SnapKV \citep{li2024snapkv}, which merges similar tokens; and compression-based methods like LongLLMLingua \citep{jiang2024longllmlingua,pan2024longllmlingua2}. Unlike these methods which primarily focus on compressing existing KV caches post-hoc, our PHD approach fundamentally rethinks the relationship between computation and KV cache by sharing KV cache across repeated tokens, maintaining the same KV cache size as the original transformer while increasing computational depth for improved performance.

\paragraph{KV Cache Management.} Beyond reduction, effective management of the KV cache is crucial. PagedAttention \citep{Kwon2023vllm} optimizes KV cache allocation and access patterns, while FlashDecoding \citep{dao2023flashdecoding} and FlashDecoding++ \citep{Hong2023flashdecoding++} enhance the efficiency of attention computations during decoding. Our \textit{PHD}-Transformer complements these methods by focusing on the efficient utilization of the KV cache through controlled sharing patterns, and could potentially be combined with these management techniques for further efficiency gains.

\paragraph{Attention Kernels and Hardware Optimization.} Specialized attention implementations like FlashAttention series \citep{dao2023flashattention,dao2024flashattention2,shah2024flashattention3} and RingAttention \citep{liu2023ringattention,brandon2023stripedattention} optimize memory access patterns to accelerate attention computation. Our \ours approach is orthogonal to these kernel optimizations and could leverage them for implementation, potentially providing compounded efficiency gains while addressing the core issue of scaling computational depth without proportional KV cache increases.

\subsection{Latent Thinking Transformers}

\paragraph{Hidden Decoding Tokens Insertion.} With the success of Chain-of-thought reasoning~\cite{wei2022chain} in the inference stage, researchers have long been interested in developing pretraining language models that can reason themselves. In the early stage of this topic, ~\citet{goyalthink} propose to insert learnable \textit{pause} tokens randomly in the pretraining sequence and observes improvement on reasoning benchmarks including GSM8K~\cite{cobbe2021training}, NaturalQ~\cite{kwiatkowski2019natural}, CommonsenseQA~\cite{talmor2018commonsenseqa}, etc. Quiet-Star~\cite{zelikman2024quiet} further proposes a reinforce framework to insert thinking tokens via online exploration. More recently, researchers~\cite{hao2024training,tack2025llm,mohtashami2023cotformer} have converted these discrete thinking tokens to continuous signals and are proven effective beyond reasoning benchmarks. However, their applicability is largely limited due to the linear increase of KV cache size. Our proposed \ours also leverages continuous signals, but introduces well-designed KV cache management to maintain an acceptable increase of KV cache, leading to applicability during inference.


\paragraph{Recurrent Latent Thinking Layers.} Another line of researches seek for recurrent approaches in reusing model parameters. Both ~\citet{chen2025inner,geiping2025scaling} propose to use recurrent transformer block or transformer layer to scale the computation of decoding tokens, either statically or adaptively. However, recurrent models cannot be used in parallel generation, leading to limited efficiency during inference.

In summary, our \textit{PHD}-Transformer series introduce a unique approach to model scaling by focusing on computational depth without proportional increases in KV cache size. By combining token repetition with efficient attention patterns specifically for \textit{hidden decoding tokens}, PHD achieves performance scaling without the prohibitive memory costs typically associated with length scaling approaches. Leveraging insights from attention pattern analysis in works like MInference \citep{jiang2024minference}, we develop a practical pretraining length scaling technique that addresses a key bottleneck in current LLM inference where models are typically memory-bound rather than compute-bound.
\section{Conclusion}

In this paper, we establish pre-training length scaling as an efficient paradigm for enhancing transformer models through our Parallel Hidden Decoding (PHD) framework.
By strategically repeating input tokens while retaining only original tokens in the KV cache, \textit{PHD}-Transformer achieves significant performance gains without increasing the KV cache size.
The \textit{PHD-SWA} variant further preserves local dependencies through sliding window attention,
while \textit{PHD-CSWA} eliminates linear pre-filling latency growth through chunk-wise sliding window attention.
Experiments demonstrate consistent improvements across multiple benchmarks, validating that length scaling can be both effective and resource-efficient during inference.

\clearpage

\bibliographystyle{plainnat}
\bibliography{main}

\begin{thebibliography}{61}
\providecommand{\natexlab}[1]{#1}
\providecommand{\url}[1]{\texttt{#1}}
\expandafter\ifx\csname urlstyle\endcsname\relax
  \providecommand{\doi}[1]{doi: #1}\else
  \providecommand{\doi}{doi: \begingroup \urlstyle{rm}\Url}\fi

\bibitem[Abdin et~al.(2024)Abdin, Jacobs, Awan, Aneja, Awadallah, Awadalla, Bach, Bahree, Bakhtiari, Behl, Benhaim, Bilenko, Bjorck, Bubeck, Cai, Mendes, Chen, Chaudhary, Chopra, Giorno, de~Rosa, Dixon, Eldan, Iter, Garg, Goswami, Gunasekar, Haider, Hao, Hewett, Huynh, Javaheripi, Jin, Kauffmann, Karampatziakis, Kim, Khademi, Kurilenko, Lee, Lee, Li, Liang, Liu, Lin, Lin, Madan, Mitra, Modi, Nguyen, Norick, Patra, Perez{-}Becker, Portet, Pryzant, Qin, Radmilac, Rosset, Roy, Ruwase, Saarikivi, Saied, Salim, Santacroce, Shah, Shang, Sharma, Song, Tanaka, Wang, Ward, Wang, Witte, Wyatt, Xu, Xu, Yadav, Yang, Yang, Yu, Zhang, Zhang, Zhang, Zhang, Zhang, Zhang, Zhang, and Zhou]{Abdin2024phi3}
Marah~I Abdin, Sam~Ade Jacobs, Ammar~Ahmad Awan, Jyoti Aneja, Ahmed Awadallah, Hany Awadalla, Nguyen Bach, Amit Bahree, Arash Bakhtiari, Harkirat~S. Behl, Alon Benhaim, Misha Bilenko, Johan Bjorck, S{\'{e}}bastien Bubeck, Martin Cai, Caio C{\'{e}}sar~Teodoro Mendes, Weizhu Chen, Vishrav Chaudhary, Parul Chopra, Allie~Del Giorno, Gustavo de~Rosa, Matthew Dixon, Ronen Eldan, Dan Iter, Amit Garg, Abhishek Goswami, Suriya Gunasekar, Emman Haider, Junheng Hao, Russell~J. Hewett, Jamie Huynh, Mojan Javaheripi, Xin Jin, Piero Kauffmann, Nikos Karampatziakis, Dongwoo Kim, Mahoud Khademi, Lev Kurilenko, James~R. Lee, Yin~Tat Lee, Yuanzhi Li, Chen Liang, Weishung Liu, Eric Lin, Zeqi Lin, Piyush Madan, Arindam Mitra, Hardik Modi, Anh Nguyen, Brandon Norick, Barun Patra, Daniel Perez{-}Becker, Thomas Portet, Reid Pryzant, Heyang Qin, Marko Radmilac, Corby Rosset, Sambudha Roy, Olatunji Ruwase, Olli Saarikivi, Amin Saied, Adil Salim, Michael Santacroce, Shital Shah, Ning Shang, Hiteshi Sharma, Xia Song, Masahiro Tanaka,
  Xin Wang, Rachel Ward, Guanhua Wang, Philipp Witte, Michael Wyatt, Can Xu, Jiahang Xu, Sonali Yadav, Fan Yang, Ziyi Yang, Donghan Yu, Chengruidong Zhang, Cyril Zhang, Jianwen Zhang, Li~Lyna Zhang, Yi~Zhang, Yue Zhang, Yunan Zhang, and Xiren Zhou.
\newblock Phi-3 technical report: {A} highly capable language model locally on your phone.
\newblock \emph{CoRR}, abs/2404.14219, 2024.
\newblock \doi{10.48550/ARXIV.2404.14219}.
\newblock URL \url{https://doi.org/10.48550/arXiv.2404.14219}.

\bibitem[Ainslie et~al.(2023)Ainslie, Lee-Thorp, de~Jong, Zemlyanskiy, Lebron, and Sanghai]{ainslie2023gqa}
Joshua Ainslie, James Lee-Thorp, Michiel de~Jong, Yury Zemlyanskiy, Federico Lebron, and Sumit Sanghai.
\newblock Gqa: Training generalized multi-query transformer models from multi-head checkpoints.
\newblock In \emph{Proceedings of the 2023 Conference on Empirical Methods in Natural Language Processing}, pages 4895--4901, 2023.

\bibitem[Beltagy et~al.(2020)Beltagy, Peters, and Cohan]{beltagy2020longformer}
Iz~Beltagy, Matthew~E. Peters, and Arman Cohan.
\newblock Longformer: The long-document transformer.
\newblock \emph{CoRR}, abs/2004.05150, 2020.
\newblock URL \url{https://arxiv.org/abs/2004.05150}.

\bibitem[Bisk et~al.(2020)Bisk, Zellers, Gao, Choi, et~al.]{bisk2020piqa}
Yonatan Bisk, Rowan Zellers, Jianfeng Gao, Yejin Choi, et~al.
\newblock Piqa: Reasoning about physical commonsense in natural language.
\newblock In \emph{Proceedings of the AAAI conference on artificial intelligence}, volume~34, pages 7432--7439, 2020.

\bibitem[Brandon et~al.(2023)Brandon, Nrusimha, Qian, Ankner, Jin, Song, and Ragan{-}Kelley]{brandon2023stripedattention}
William Brandon, Aniruddha Nrusimha, Kevin Qian, Zachary Ankner, Tian Jin, Zhiye Song, and Jonathan Ragan{-}Kelley.
\newblock Striped attention: Faster ring attention for causal transformers.
\newblock \emph{CoRR}, abs/2311.09431, 2023.
\newblock \doi{10.48550/ARXIV.2311.09431}.
\newblock URL \url{https://doi.org/10.48550/arXiv.2311.09431}.

\bibitem[Chen et~al.(2024)Chen, Liao, Li, and Fan]{chen2024step}
Guoxin Chen, Minpeng Liao, Chengxi Li, and Kai Fan.
\newblock Step-level value preference optimization for mathematical reasoning.
\newblock \emph{arXiv preprint arXiv:2406.10858}, 2024.

\bibitem[Chen et~al.(2025)Chen, Shang, Zhang, Xie, Sheng, Liu, Wang, Sun, Wu, and Wang]{chen2025inner}
Yilong Chen, Junyuan Shang, Zhenyu Zhang, Yanxi Xie, Jiawei Sheng, Tingwen Liu, Shuohuan Wang, Yu~Sun, Hua Wu, and Haifeng Wang.
\newblock Inner thinking transformer: Leveraging dynamic depth scaling to foster adaptive internal thinking.
\newblock \emph{arXiv preprint arXiv:2502.13842}, 2025.

\bibitem[Child et~al.(2019)Child, Gray, Radford, and Sutskever]{child2019sparsetransformers}
Rewon Child, Scott Gray, Alec Radford, and Ilya Sutskever.
\newblock Generating long sequences with sparse transformers.
\newblock \emph{CoRR}, abs/1904.10509, 2019.
\newblock URL \url{http://arxiv.org/abs/1904.10509}.

\bibitem[Clark et~al.(2022)Clark, de~Las~Casas, Guy, Mensch, Paganini, Hoffmann, Damoc, Hechtman, Cai, Borgeaud, et~al.]{clark2022unified}
Aidan Clark, Diego de~Las~Casas, Aurelia Guy, Arthur Mensch, Michela Paganini, Jordan Hoffmann, Bogdan Damoc, Blake Hechtman, Trevor Cai, Sebastian Borgeaud, et~al.
\newblock Unified scaling laws for routed language models.
\newblock In \emph{International conference on machine learning}, pages 4057--4086. PMLR, 2022.

\bibitem[Clark et~al.(2018)Clark, Cowhey, Etzioni, Khot, Sabharwal, Schoenick, and Tafjord]{clark2018arc}
Peter Clark, Isaac Cowhey, Oren Etzioni, Tushar Khot, Ashish Sabharwal, Carissa Schoenick, and Oyvind Tafjord.
\newblock Think you have solved question answering? try arc, the ai2 reasoning challenge.
\newblock \emph{arXiv preprint arXiv:1803.05457}, 2018.

\bibitem[Cobbe et~al.(2021)Cobbe, Kosaraju, Bavarian, Chen, Jun, Kaiser, Plappert, Tworek, Hilton, Nakano, et~al.]{cobbe2021training}
Karl Cobbe, Vineet Kosaraju, Mohammad Bavarian, Mark Chen, Heewoo Jun, Lukasz Kaiser, Matthias Plappert, Jerry Tworek, Jacob Hilton, Reiichiro Nakano, et~al.
\newblock Training verifiers to solve math word problems.
\newblock \emph{arXiv preprint arXiv:2110.14168}, 2021.

\bibitem[Dao(2024)]{dao2024flashattention2}
Tri Dao.
\newblock Flashattention-2: Faster attention with better parallelism and work partitioning.
\newblock In \emph{The Twelfth International Conference on Learning Representations, {ICLR} 2024, Vienna, Austria, May 7-11, 2024}. OpenReview.net, 2024.
\newblock URL \url{https://openreview.net/forum?id=mZn2Xyh9Ec}.

\bibitem[Dao et~al.(2022)Dao, Fu, Ermon, Rudra, and R{\'{e}}]{dao2023flashattention}
Tri Dao, Daniel~Y. Fu, Stefano Ermon, Atri Rudra, and Christopher R{\'{e}}.
\newblock Flashattention: Fast and memory-efficient exact attention with io-awareness.
\newblock In Sanmi Koyejo, S.~Mohamed, A.~Agarwal, Danielle Belgrave, K.~Cho, and A.~Oh, editors, \emph{Advances in Neural Information Processing Systems 35: Annual Conference on Neural Information Processing Systems 2022, NeurIPS 2022, New Orleans, LA, USA, November 28 - December 9, 2022}, 2022.
\newblock URL \url{http://papers.nips.cc/paper\_files/paper/2022/hash/67d57c32e20fd0a7a302cb81d36e40d5-Abstract-Conference.html}.

\bibitem[Dao et~al.(2023)Dao, Haziza, Massa, and Sizov]{dao2023flashdecoding}
Tri Dao, Daniel Haziza, Francisco Massa, and Grigory Sizov.
\newblock Flash-decoding for long-context inference, October 2023.
\newblock URL \url{https://crfm.stanford.edu/2023/10/12/flashdecoding.html}.
\newblock Accessed: 2024-9-29.

\bibitem[Ding et~al.(2023)Ding, Ma, Dong, Zhang, Huang, Wang, Zheng, and Wei]{ding2023longnet}
Jiayu Ding, Shuming Ma, Li~Dong, Xingxing Zhang, Shaohan Huang, Wenhui Wang, Nanning Zheng, and Furu Wei.
\newblock Longnet: Scaling transformers to 1, 000, 000, 000 tokens.
\newblock \emph{CoRR}, abs/2307.02486, 2023.
\newblock \doi{10.48550/ARXIV.2307.02486}.
\newblock URL \url{https://doi.org/10.48550/arXiv.2307.02486}.

\bibitem[Geiping et~al.(2025)Geiping, McLeish, Jain, Kirchenbauer, Singh, Bartoldson, Kailkhura, Bhatele, and Goldstein]{geiping2025scaling}
Jonas Geiping, Sean McLeish, Neel Jain, John Kirchenbauer, Siddharth Singh, Brian~R Bartoldson, Bhavya Kailkhura, Abhinav Bhatele, and Tom Goldstein.
\newblock Scaling up test-time compute with latent reasoning: A recurrent depth approach.
\newblock \emph{arXiv preprint arXiv:2502.05171}, 2025.

\bibitem[Goyal et~al.()Goyal, Ji, Rawat, Menon, Kumar, and Nagarajan]{goyalthink}
Sachin Goyal, Ziwei Ji, Ankit~Singh Rawat, Aditya~Krishna Menon, Sanjiv Kumar, and Vaishnavh Nagarajan.
\newblock Think before you speak: Training language models with pause tokens.
\newblock In \emph{The Twelfth International Conference on Learning Representations}.

\bibitem[Guo et~al.(2025)Guo, Yang, Zhang, Song, Zhang, Xu, Zhu, Ma, Wang, Bi, et~al.]{guo2025deepseek}
Daya Guo, Dejian Yang, Haowei Zhang, Junxiao Song, Ruoyu Zhang, Runxin Xu, Qihao Zhu, Shirong Ma, Peiyi Wang, Xiao Bi, et~al.
\newblock Deepseek-r1: Incentivizing reasoning capability in llms via reinforcement learning.
\newblock \emph{arXiv preprint arXiv:2501.12948}, 2025.

\bibitem[Han et~al.(2024)Han, Koh, Seo, Chang, and Sohn]{han2024psydial}
Ji-Eun Han, Jun-Seok Koh, Hyeon-Tae Seo, Du-Seong Chang, and Kyung-Ah Sohn.
\newblock Psydial: personality-based synthetic dialogue generation using large language models.
\newblock \emph{arXiv preprint arXiv:2404.00930}, 2024.

\bibitem[Hao et~al.(2024)Hao, Sukhbaatar, Su, Li, Hu, Weston, and Tian]{hao2024training}
Shibo Hao, Sainbayar Sukhbaatar, DiJia Su, Xian Li, Zhiting Hu, Jason Weston, and Yuandong Tian.
\newblock Training large language models to reason in a continuous latent space.
\newblock \emph{arXiv preprint arXiv:2412.06769}, 2024.

\bibitem[Hendrycks et~al.(2020)Hendrycks, Burns, Basart, Zou, Mazeika, Song, and Steinhardt]{hendrycks2020measuring}
Dan Hendrycks, Collin Burns, Steven Basart, Andy Zou, Mantas Mazeika, Dawn Song, and Jacob Steinhardt.
\newblock Measuring massive multitask language understanding.
\newblock \emph{arXiv preprint arXiv:2009.03300}, 2020.

\bibitem[Hernandez et~al.(2021)Hernandez, Kaplan, Henighan, and McCandlish]{hernandez2021scaling}
Danny Hernandez, Jared Kaplan, Tom Henighan, and Sam McCandlish.
\newblock Scaling laws for transfer.
\newblock \emph{arXiv preprint arXiv:2102.01293}, 2021.

\bibitem[Hong et~al.(2023)Hong, Dai, Xu, Mao, Li, Liu, Chen, Dong, and Wang]{Hong2023flashdecoding++}
Ke~Hong, Guohao Dai, Jiaming Xu, Qiuli Mao, Xiuhong Li, Jun Liu, Kangdi Chen, Yuhan Dong, and Yu~Wang.
\newblock Flashdecoding++: Faster large language model inference on gpus.
\newblock \emph{CoRR}, abs/2311.01282, 2023.
\newblock \doi{10.48550/ARXIV.2311.01282}.
\newblock URL \url{https://doi.org/10.48550/arXiv.2311.01282}.

\bibitem[Jiang et~al.(2023)Jiang, Sablayrolles, Mensch, Bamford, Chaplot, de~Las~Casas, Bressand, Lengyel, Lample, Saulnier, Lavaud, Lachaux, Stock, Scao, Lavril, Wang, Lacroix, and Sayed]{jiang2023mistral7b}
Albert~Q. Jiang, Alexandre Sablayrolles, Arthur Mensch, Chris Bamford, Devendra~Singh Chaplot, Diego de~Las~Casas, Florian Bressand, Gianna Lengyel, Guillaume Lample, Lucile Saulnier, L{\'{e}}lio~Renard Lavaud, Marie{-}Anne Lachaux, Pierre Stock, Teven~Le Scao, Thibaut Lavril, Thomas Wang, Timoth{\'{e}}e Lacroix, and William~El Sayed.
\newblock Mistral 7b.
\newblock \emph{CoRR}, abs/2310.06825, 2023.
\newblock \doi{10.48550/ARXIV.2310.06825}.
\newblock URL \url{https://doi.org/10.48550/arXiv.2310.06825}.

\bibitem[Jiang et~al.(2024{\natexlab{a}})Jiang, Li, Zhang, Wu, Luo, Ahn, Han, Abdi, Li, Lin, et~al.]{jiang2024minference}
Huiqiang Jiang, Yucheng Li, Chengruidong Zhang, Qianhui Wu, Xufang Luo, Surin Ahn, Zhenhua Han, Amir Abdi, Dongsheng Li, Chin-Yew Lin, et~al.
\newblock Minference 1.0: Accelerating pre-filling for long-context llms via dynamic sparse attention.
\newblock \emph{Advances in Neural Information Processing Systems}, 37:\penalty0 52481--52515, 2024{\natexlab{a}}.

\bibitem[Jiang et~al.(2024{\natexlab{b}})Jiang, Wu, Luo, Li, Lin, Yang, and Qiu]{jiang2024longllmlingua}
Huiqiang Jiang, Qianhui Wu, Xufang Luo, Dongsheng Li, Chin{-}Yew Lin, Yuqing Yang, and Lili Qiu.
\newblock Longllmlingua: Accelerating and enhancing llms in long context scenarios via prompt compression.
\newblock In Lun{-}Wei Ku, Andre Martins, and Vivek Srikumar, editors, \emph{Proceedings of the 62nd Annual Meeting of the Association for Computational Linguistics (Volume 1: Long Papers), {ACL} 2024, Bangkok, Thailand, August 11-16, 2024}, pages 1658--1677. Association for Computational Linguistics, 2024{\natexlab{b}}.
\newblock \doi{10.18653/V1/2024.ACL-LONG.91}.
\newblock URL \url{https://doi.org/10.18653/v1/2024.acl-long.91}.

\bibitem[Jimenez et~al.(2023)Jimenez, Yang, Wettig, Yao, Pei, Press, and Narasimhan]{jimenez2023swe}
Carlos~E Jimenez, John Yang, Alexander Wettig, Shunyu Yao, Kexin Pei, Ofir Press, and Karthik Narasimhan.
\newblock Swe-bench: Can language models resolve real-world github issues?
\newblock \emph{arXiv preprint arXiv:2310.06770}, 2023.

\bibitem[Kaplan et~al.(2020)Kaplan, McCandlish, Henighan, Brown, Chess, Child, Gray, Radford, Wu, and Amodei]{kaplan2020scaling}
Jared Kaplan, Sam McCandlish, Tom Henighan, Tom~B Brown, Benjamin Chess, Rewon Child, Scott Gray, Alec Radford, Jeffrey Wu, and Dario Amodei.
\newblock Scaling laws for neural language models.
\newblock \emph{arXiv preprint arXiv:2001.08361}, 2020.

\bibitem[Kwiatkowski et~al.(2019)Kwiatkowski, Palomaki, Redfield, Collins, Parikh, Alberti, Epstein, Polosukhin, Devlin, Lee, et~al.]{kwiatkowski2019natural}
Tom Kwiatkowski, Jennimaria Palomaki, Olivia Redfield, Michael Collins, Ankur Parikh, Chris Alberti, Danielle Epstein, Illia Polosukhin, Jacob Devlin, Kenton Lee, et~al.
\newblock Natural questions: a benchmark for question answering research.
\newblock \emph{Transactions of the Association for Computational Linguistics}, 7:\penalty0 453--466, 2019.

\bibitem[Kwon et~al.(2023)Kwon, Li, Zhuang, Sheng, Zheng, Yu, Gonzalez, Zhang, and Stoica]{Kwon2023vllm}
Woosuk Kwon, Zhuohan Li, Siyuan Zhuang, Ying Sheng, Lianmin Zheng, Cody~Hao Yu, Joseph Gonzalez, Hao Zhang, and Ion Stoica.
\newblock Efficient memory management for large language model serving with pagedattention.
\newblock In Jason Flinn, Margo~I. Seltzer, Peter Druschel, Antoine Kaufmann, and Jonathan Mace, editors, \emph{Proceedings of the 29th Symposium on Operating Systems Principles, {SOSP} 2023, Koblenz, Germany, October 23-26, 2023}, pages 611--626. {ACM}, 2023.
\newblock \doi{10.1145/3600006.3613165}.
\newblock URL \url{https://doi.org/10.1145/3600006.3613165}.

\bibitem[Li et~al.(2025)Li, Gong, Yang, Shan, Liu, Zhu, Zhang, Guo, Chen, Li, et~al.]{li2025minimax}
Aonian Li, Bangwei Gong, Bo~Yang, Boji Shan, Chang Liu, Cheng Zhu, Chunhao Zhang, Congchao Guo, Da~Chen, Dong Li, et~al.
\newblock Minimax-01: Scaling foundation models with lightning attention.
\newblock \emph{arXiv preprint arXiv:2501.08313}, 2025.

\bibitem[Li et~al.(2024)Li, Huang, Yang, Venkitesh, Locatelli, Ye, Cai, Lewis, and Chen]{li2024snapkv}
Yuhong Li, Yingbing Huang, Bowen Yang, Bharat Venkitesh, Acyr Locatelli, Hanchen Ye, Tianle Cai, Patrick Lewis, and Deming Chen.
\newblock Snapkv: {LLM} knows what you are looking for before generation.
\newblock \emph{CoRR}, abs/2404.14469, 2024.
\newblock \doi{10.48550/ARXIV.2404.14469}.
\newblock URL \url{https://doi.org/10.48550/arXiv.2404.14469}.

\bibitem[Liu et~al.(2024{\natexlab{a}})Liu, Feng, Wang, Wang, Liu, Zhao, Dengr, Ruan, Dai, Guo, et~al.]{liu2024deepseek}
Aixin Liu, Bei Feng, Bin Wang, Bingxuan Wang, Bo~Liu, Chenggang Zhao, Chengqi Dengr, Chong Ruan, Damai Dai, Daya Guo, et~al.
\newblock Deepseek-v2: A strong, economical, and efficient mixture-of-experts language model.
\newblock \emph{arXiv preprint arXiv:2405.04434}, 2024{\natexlab{a}}.

\bibitem[Liu et~al.(2024{\natexlab{b}})Liu, Feng, Xue, Wang, Wu, Lu, Zhao, Deng, Zhang, Ruan, et~al.]{liu2024deepseekv3}
Aixin Liu, Bei Feng, Bing Xue, Bingxuan Wang, Bochao Wu, Chengda Lu, Chenggang Zhao, Chengqi Deng, Chenyu Zhang, Chong Ruan, et~al.
\newblock Deepseek-v3 technical report.
\newblock \emph{arXiv preprint arXiv:2412.19437}, 2024{\natexlab{b}}.

\bibitem[Liu et~al.(2023)Liu, Zaharia, and Abbeel]{liu2023ringattention}
Hao Liu, Matei Zaharia, and Pieter Abbeel.
\newblock Ring attention with blockwise transformers for near-infinite context.
\newblock \emph{CoRR}, abs/2310.01889, 2023.
\newblock \doi{10.48550/ARXIV.2310.01889}.
\newblock URL \url{https://doi.org/10.48550/arXiv.2310.01889}.

\bibitem[Mohtashami et~al.(2023)Mohtashami, Pagliardini, and Jaggi]{mohtashami2023cotformer}
Amirkeivan Mohtashami, Matteo Pagliardini, and Martin Jaggi.
\newblock Cotformer: More tokens with attention make up for less depth.
\newblock In \emph{Workshop on Advancing Neural Network Training: Computational Efficiency, Scalability, and Resource Optimization (WANT@ NeurIPS 2023)}, 2023.

\bibitem[OLMo et~al.(2024)OLMo, Walsh, Soldaini, Groeneveld, Lo, Arora, Bhagia, Gu, Huang, Jordan, et~al.]{olmo20242}
Team OLMo, Pete Walsh, Luca Soldaini, Dirk Groeneveld, Kyle Lo, Shane Arora, Akshita Bhagia, Yuling Gu, Shengyi Huang, Matt Jordan, et~al.
\newblock 2 olmo 2 furious.
\newblock \emph{arXiv preprint arXiv:2501.00656}, 2024.

\bibitem[OpenAI(2024)]{o1}
OpenAI.
\newblock Learning to reason with llms, 2024.
\newblock URL \url{https://openai.com/index/learning-to-reason-with-llms/}.

\bibitem[OpenAI(2025)]{o3}
OpenAI.
\newblock Learning to reason with llms, 2025.
\newblock URL \url{https://openai.com/index/openai-o3-mini/}.

\bibitem[Pan et~al.(2024)Pan, Wu, Jiang, Xia, Luo, Zhang, Lin, R{\"{u}}hle, Yang, Lin, Zhao, Qiu, and Zhang]{pan2024longllmlingua2}
Zhuoshi Pan, Qianhui Wu, Huiqiang Jiang, Menglin Xia, Xufang Luo, Jue Zhang, Qingwei Lin, Victor R{\"{u}}hle, Yuqing Yang, Chin{-}Yew Lin, H.~Vicky Zhao, Lili Qiu, and Dongmei Zhang.
\newblock Llmlingua-2: Data distillation for efficient and faithful task-agnostic prompt compression.
\newblock In Lun{-}Wei Ku, Andre Martins, and Vivek Srikumar, editors, \emph{Findings of the Association for Computational Linguistics, {ACL} 2024, Bangkok, Thailand and virtual meeting, August 11-16, 2024}, pages 963--981. Association for Computational Linguistics, 2024.
\newblock \doi{10.18653/V1/2024.FINDINGS-ACL.57}.
\newblock URL \url{https://doi.org/10.18653/v1/2024.findings-acl.57}.

\bibitem[Rein et~al.(2024)Rein, Hou, Stickland, Petty, Pang, Dirani, Michael, and Bowman]{rein2024gpqa}
David Rein, Betty~Li Hou, Asa~Cooper Stickland, Jackson Petty, Richard~Yuanzhe Pang, Julien Dirani, Julian Michael, and Samuel~R Bowman.
\newblock Gpqa: A graduate-level google-proof q\&a benchmark.
\newblock In \emph{First Conference on Language Modeling}, 2024.

\bibitem[Ribar et~al.(2023)Ribar, Chelombiev, Hudlass-Galley, Blake, Luschi, and Orr]{ribar2023sparq}
Luka Ribar, Ivan Chelombiev, Luke Hudlass-Galley, Charlie Blake, Carlo Luschi, and Douglas Orr.
\newblock Sparq attention: Bandwidth-efficient llm inference.
\newblock \emph{arXiv preprint arXiv:2312.04985}, 2023.

\bibitem[Sakaguchi et~al.(2021)Sakaguchi, Bras, Bhagavatula, and Choi]{sakaguchi2021winogrande}
Keisuke Sakaguchi, Ronan~Le Bras, Chandra Bhagavatula, and Yejin Choi.
\newblock Winogrande: An adversarial winograd schema challenge at scale.
\newblock \emph{Communications of the ACM}, 64\penalty0 (9):\penalty0 99--106, 2021.

\bibitem[Schulman et~al.(2017)Schulman, Wolski, Dhariwal, Radford, and Klimov]{schulman2017proximal}
John Schulman, Filip Wolski, Prafulla Dhariwal, Alec Radford, and Oleg Klimov.
\newblock Proximal policy optimization algorithms.
\newblock \emph{arXiv preprint arXiv:1707.06347}, 2017.

\bibitem[Shah et~al.(2024)Shah, Bikshandi, Zhang, Thakkar, Ramani, and Dao]{shah2024flashattention3}
Jay Shah, Ganesh Bikshandi, Ying Zhang, Vijay Thakkar, Pradeep Ramani, and Tri Dao.
\newblock Flashattention-3: Fast and accurate attention with asynchrony and low-precision.
\newblock \emph{CoRR}, abs/2407.08608, 2024.
\newblock \doi{10.48550/ARXIV.2407.08608}.
\newblock URL \url{https://doi.org/10.48550/arXiv.2407.08608}.

\bibitem[Shao et~al.(2024)Shao, Wang, Zhu, Xu, Song, Bi, Zhang, Zhang, Li, Wu, et~al.]{shao2024deepseekmath}
Zhihong Shao, Peiyi Wang, Qihao Zhu, Runxin Xu, Junxiao Song, Xiao Bi, Haowei Zhang, Mingchuan Zhang, YK~Li, Y~Wu, et~al.
\newblock Deepseekmath: Pushing the limits of mathematical reasoning in open language models.
\newblock \emph{arXiv preprint arXiv:2402.03300}, 2024.

\bibitem[Shi et~al.(2021)Shi, Gao, Ren, Xu, Liang, Li, and Kwok]{shi2021sparsebert}
Han Shi, Jiahui Gao, Xiaozhe Ren, Hang Xu, Xiaodan Liang, Zhenguo Li, and James~Tin{-}Yau Kwok.
\newblock Sparsebert: Rethinking the importance analysis in self-attention.
\newblock In Marina Meila and Tong Zhang, editors, \emph{Proceedings of the 38th International Conference on Machine Learning, {ICML} 2021, 18-24 July 2021, Virtual Event}, volume 139 of \emph{Proceedings of Machine Learning Research}, pages 9547--9557. {PMLR}, 2021.
\newblock URL \url{http://proceedings.mlr.press/v139/shi21a.html}.

\bibitem[Tack et~al.(2025)Tack, Lanchantin, Yu, Cohen, Kulikov, Lan, Hao, Tian, Weston, and Li]{tack2025llm}
Jihoon Tack, Jack Lanchantin, Jane Yu, Andrew Cohen, Ilia Kulikov, Janice Lan, Shibo Hao, Yuandong Tian, Jason Weston, and Xian Li.
\newblock Llm pretraining with continuous concepts.
\newblock \emph{arXiv preprint arXiv:2502.08524}, 2025.

\bibitem[Talmor et~al.(2018)Talmor, Herzig, Lourie, and Berant]{talmor2018commonsenseqa}
Alon Talmor, Jonathan Herzig, Nicholas Lourie, and Jonathan Berant.
\newblock Commonsenseqa: A question answering challenge targeting commonsense knowledge.
\newblock \emph{arXiv preprint arXiv:1811.00937}, 2018.

\bibitem[Tang et~al.(2024)Tang, Zhao, Zhu, Xiao, Kasikci, and Han]{tang2024quest}
Jiaming Tang, Yilong Zhao, Kan Zhu, Guangxuan Xiao, Baris Kasikci, and Song Han.
\newblock Quest: Query-aware sparsity for efficient long-context llm inference.
\newblock In \emph{International Conference on Machine Learning}, pages 47901--47911. PMLR, 2024.

\bibitem[Team et~al.(2023)Team, Anil, Borgeaud, Alayrac, Yu, Soricut, Schalkwyk, Dai, Hauth, Millican, et~al.]{team2023gemini}
Gemini Team, Rohan Anil, Sebastian Borgeaud, Jean-Baptiste Alayrac, Jiahui Yu, Radu Soricut, Johan Schalkwyk, Andrew~M Dai, Anja Hauth, Katie Millican, et~al.
\newblock Gemini: a family of highly capable multimodal models.
\newblock \emph{arXiv preprint arXiv:2312.11805}, 2023.

\bibitem[Team et~al.(2024)Team, Georgiev, Lei, Burnell, Bai, Gulati, Tanzer, Vincent, Pan, Wang, et~al.]{team2024gemini}
Gemini Team, Petko Georgiev, Ving~Ian Lei, Ryan Burnell, Libin Bai, Anmol Gulati, Garrett Tanzer, Damien Vincent, Zhufeng Pan, Shibo Wang, et~al.
\newblock Gemini 1.5: Unlocking multimodal understanding across millions of tokens of context.
\newblock \emph{arXiv preprint arXiv:2403.05530}, 2024.

\bibitem[Wang et~al.(2021)Wang, Zhang, and Han]{wang2021spatten}
Hanrui Wang, Zhekai Zhang, and Song Han.
\newblock Spatten: Efficient sparse attention architecture with cascade token and head pruning.
\newblock In \emph{{IEEE} International Symposium on High-Performance Computer Architecture, {HPCA} 2021, Seoul, South Korea, February 27 - March 3, 2021}, pages 97--110. {IEEE}, 2021.
\newblock \doi{10.1109/HPCA51647.2021.00018}.
\newblock URL \url{https://doi.org/10.1109/HPCA51647.2021.00018}.

\bibitem[Wang et~al.(2024)Wang, Li, Song, Xu, Tang, Zhuge, Pan, Song, Li, Singh, et~al.]{wang2024openhands}
Xingyao Wang, Boxuan Li, Yufan Song, Frank~F Xu, Xiangru Tang, Mingchen Zhuge, Jiayi Pan, Yueqi Song, Bowen Li, Jaskirat Singh, et~al.
\newblock Openhands: An open platform for ai software developers as generalist agents.
\newblock In \emph{The Thirteenth International Conference on Learning Representations}, 2024.

\bibitem[Wei et~al.(2022)Wei, Wang, Schuurmans, Bosma, Xia, Chi, Le, Zhou, et~al.]{wei2022chain}
Jason Wei, Xuezhi Wang, Dale Schuurmans, Maarten Bosma, Fei Xia, Ed~Chi, Quoc~V Le, Denny Zhou, et~al.
\newblock Chain-of-thought prompting elicits reasoning in large language models.
\newblock \emph{Advances in neural information processing systems}, 35:\penalty0 24824--24837, 2022.

\bibitem[Xiao et~al.(2024)Xiao, Tian, Chen, Han, and Lewis]{xiao2024streamingllm}
Guangxuan Xiao, Yuandong Tian, Beidi Chen, Song Han, and Mike Lewis.
\newblock Efficient streaming language models with attention sinks.
\newblock In \emph{The Twelfth International Conference on Learning Representations, {ICLR} 2024, Vienna, Austria, May 7-11, 2024}. OpenReview.net, 2024.
\newblock URL \url{https://openreview.net/forum?id=NG7sS51zVF}.

\bibitem[Yuan et~al.(2025)Yuan, Gao, Dai, Luo, Zhao, Zhang, Xie, Wei, Wang, Xiao, et~al.]{yuan2025native}
Jingyang Yuan, Huazuo Gao, Damai Dai, Junyu Luo, Liang Zhao, Zhengyan Zhang, Zhenda Xie, YX~Wei, Lean Wang, Zhiping Xiao, et~al.
\newblock Native sparse attention: Hardware-aligned and natively trainable sparse attention.
\newblock \emph{arXiv preprint arXiv:2502.11089}, 2025.

\bibitem[Zaheer et~al.(2020)Zaheer, Guruganesh, Dubey, Ainslie, Alberti, Onta{\~{n}}{\'{o}}n, Pham, Ravula, Wang, Yang, and Ahmed]{zaheer2020bigbird}
Manzil Zaheer, Guru Guruganesh, Kumar~Avinava Dubey, Joshua Ainslie, Chris Alberti, Santiago Onta{\~{n}}{\'{o}}n, Philip Pham, Anirudh Ravula, Qifan Wang, Li~Yang, and Amr Ahmed.
\newblock Big bird: Transformers for longer sequences.
\newblock In Hugo Larochelle, Marc'Aurelio Ranzato, Raia Hadsell, Maria{-}Florina Balcan, and Hsuan{-}Tien Lin, editors, \emph{Advances in Neural Information Processing Systems 33: Annual Conference on Neural Information Processing Systems 2020, NeurIPS 2020, December 6-12, 2020, virtual}, 2020.
\newblock URL \url{https://proceedings.neurips.cc/paper/2020/hash/c8512d142a2d849725f31a9a7a361ab9-Abstract.html}.

\bibitem[Zelikman et~al.(2024)Zelikman, Harik, Shao, Jayasiri, Haber, and Goodman]{zelikman2024quiet}
Eric Zelikman, Georges~Raif Harik, Yijia Shao, Varuna Jayasiri, Nick Haber, and Noah Goodman.
\newblock Quiet-star: Language models can teach themselves to think before speaking.
\newblock In \emph{First Conference on Language Modeling}, 2024.

\bibitem[Zellers et~al.(2019)Zellers, Holtzman, Bisk, Farhadi, and Choi]{zellers2019hellaswag}
Rowan Zellers, Ari Holtzman, Yonatan Bisk, Ali Farhadi, and Yejin Choi.
\newblock Hellaswag: Can a machine really finish your sentence?
\newblock \emph{arXiv preprint arXiv:1905.07830}, 2019.

\bibitem[Zhang et~al.(2023)Zhang, Sheng, Zhou, Chen, Zheng, Cai, Song, Tian, R{\'e}, Barrett, et~al.]{zhang2023h2o}
Zhenyu Zhang, Ying Sheng, Tianyi Zhou, Tianlong Chen, Lianmin Zheng, Ruisi Cai, Zhao Song, Yuandong Tian, Christopher R{\'e}, Clark Barrett, et~al.
\newblock H2o: Heavy-hitter oracle for efficient generative inference of large language models.
\newblock \emph{Advances in Neural Information Processing Systems}, 36:\penalty0 34661--34710, 2023.

\end{thebibliography}

\clearpage



\end{document}